\documentclass[journal]{IEEEtran}
\usepackage{amsmath,amsfonts}
\usepackage{algorithmic}
\usepackage{array}
\usepackage{textcomp}
\usepackage{stfloats}
\usepackage{url}
\usepackage{verbatim}
\usepackage{graphicx}
\usepackage{subfigure}
\usepackage{float}
\usepackage{tikz}

\DeclareRobustCommand{\vertcircle}{
\hspace{-0.65em}
\begin{tikzpicture}[baseline=-0.7ex, scale=0.5]
    \draw (0,0) circle (0.25cm);
    \draw (-0.1,0.15) -- (-0.1,-0.15);
    \draw (0.1,0.15) -- (0.1,-0.15);
\end{tikzpicture}\hspace{-0.2em}
}

\DeclareRobustCommand{\vertcir}{
\hspace{0.2em}
\begin{tikzpicture}[baseline=(vertcir.base)]
    \node[inner sep=0pt, outer sep=0pt] (vertcir) at (0,0) {
        \begin{tikzpicture}[baseline=-0.6ex, scale=0.5]
            \draw (0,0) circle (0.25cm);
            \draw (-0.1,0.15) -- (-0.1,-0.15);
            \draw (0.1,0.15) -- (0.1,-0.15);
        \end{tikzpicture}
    };
\end{tikzpicture}
\hspace{0.2em}
}

\usepackage{xcolor}
\usepackage{colortbl}
\usepackage{multirow}
\usepackage{enumitem}
\usepackage{hyperref}
\hypersetup{
    colorlinks=true,
    linkcolor=blue,
    filecolor=blue,
    citecolor=green,
    urlcolor=blue,
    }

\usepackage{orcidlink}

\def\BibTeX{{\rm B\kern-.05em{\sc i\kern-.025em b}\kern-.08em
    T\kern-.1667em\lower.7ex\hbox{E}\kern-.125emX}}
\usepackage{balance}
\begin{document}
\title{Geometric Features Enhanced Human-Object Interaction Detection}
\author{Manli Zhu$^{\orcidlink{0000-0002-8231-5342}}$, Edmond S. L. Ho$^{\orcidlink{0000-0001-5862-106X}}$, Shuang Chen$^{\orcidlink{0000-0002-6879-7285}}$, Longzhi Yang$^{\orcidlink{0000-0003-2115-4909}}$, \textit{Senior Member, IEEE}, Hubert P. H. Shum$^{\orcidlink{0000-0001-5651-6039}\dag}$, \textit{Senior Member, IEEE}
\thanks{M. Zhu and L. Yang are with the Department of Computer and Information Sciences, Northumbria University, Newcastle upon Tyne, UK. Emails: \{manli.zhu, longzhi.yang\}@northumbria.ac.uk}
\thanks{E. S. L. Ho is with the School of Computing Science, University of Glasgow, Glasgow, UK. Email: shu-lim.ho@glasgow.ac.uk}
\thanks{S. Chen and H. P. H. Shum are with the Department of Computer Science, Durham University, Durham, UK. Emails: \{shuang.chen, hubert.shum\}@durham.ac.uk}
\thanks{$^{\dag}$Corresponding author: H. P. H. Shum}
}


\markboth{IEEE TRANSACTIONS ON INSTRUMENTATION AND MEASUREMENT}
{}

\maketitle

\begin{abstract}
Cameras are essential vision instruments to capture images for pattern detection and measurement. Human-object interaction (HOI) detection is one of the most popular pattern detection approaches for captured human-centric visual scenes. Recently, Transformer-based models have become the dominant approach for HOI detection due to their advanced network architectures and thus promising results. However, most of them follow the one-stage design of vanilla Transformer, leaving rich geometric priors under-exploited and leading to compromised performance especially when occlusion occurs. Given that geometric features tend to outperform visual ones in occluded scenarios and offer information that complements visual cues, we propose a novel end-to-end Transformer-style HOI detection model, i.e., geometric features enhanced HOI detector (GeoHOI). One key part of the model is a new unified self-supervised keypoint learning method named UniPointNet that bridges the gap of consistent keypoint representation across diverse object categories, including humans. GeoHOI effectively upgrades a Transformer-based HOI detector benefiting from the keypoints similarities measuring the likelihood of human-object interactions as well as local keypoint patches to enhance interaction query representation, so as to boost HOI predictions. Extensive experiments show that the proposed method outperforms the state-of-the-art models on V-COCO and achieves competitive performance on HICO-DET. Case study results on the post-disaster rescue with vision-based instruments showcase the applicability of the proposed GeoHOI in real-world applications. 
\end{abstract}


\section{Introduction}
Cameras, as predominant vision instruments, are extensively employed in methods that rely on visual measurements \cite{camera-vision}, such as human and object pose estimation \cite{human-pose2,pose,object-pose}. Human-object interaction (HOI) detection is one of the most popular pattern detection approaches for captured human-centric visual scenes. It involves identifying and localizing interactive human-object pairs while predicting the specific interactions between them within an image, yielding HOI triplets $\left<human, interaction, object \right>$. It plays an important role in numerous applications, such as action recognition \cite{action} and surveillance event detection \cite{surveillance1,surveillance2}.

The existing HOI detection methods generally fall into two-stage or end-to-end approaches. Two-stage approaches \cite{PMFNet,VSGNet,SGCN4HOI} typically take advantage of off-the-shelf object detectors like Fast R-CNN \cite{FastRCNN}. They first detect all instances (i.e., humans and objects) in an image, and then the interaction classification is carried out on every human-object pair. These methods may lead to sub-optimal HOI detections due to the independent optimization of two sub-problems \cite{HOITransformer}, i.e., object detection and interaction classification. In contrast, end-to-end approaches detect the components of an HOI triplet all at once \cite{SSRT}. In the earlier end-to-end attempts \cite{IPNet,PPDM}, interaction points and object proposals are detected simultaneously. The interactions are then associated with each human-object pair. However, in still images of complex scenes, such as crowded areas with interaction points overlapping among different human-object pairs, these methods could lead to inaccuracies and misinterpretations \cite{qpic,SSRT}. 

End-to-end Transformer-based models \cite{HOTR,HOITransformer,qpic} have been proposed to overcome these limitations, achieving state-of-the-art performance. Inspired by the Transformer object detector DETR \cite{detr}, these approaches frame the HOI detection as a set prediction problem, using a bipartite matching loss to align interaction queries with ground-truth HOI triplets. While successful, rich prior knowledge (e.g., the semantic features and structure information) is under-exploited due to the random initialization of parametric interaction queries. To address this limitation, \cite{SSRT,stip,cql} explored semantics, spatial features, and structure information. Nevertheless, the spatial features including instance bounding boxes and human-object layout employed in these works are too coarse to capture fine-grained relationships between human body parts and object parts. The fine-grained geometric features, such as human pose and object structure have proven to be highly effective in two-stage methods \cite{PMFNet,SGCN4HOI,cross-person-part}. However, they remain under-explored in existing Transformers due to their one-stage paradigm of HOI detection. In this work, we investigate how to enrich HOI representations with fine-grained geometric features in an end-to-end Transformer framework.

To this end, we propose a \textbf{Geo}metric features enhanced \textbf{H}uman-\textbf{O}bject \textbf{I}nteraction detection model (GeoHOI). Given that geometric features tend to outperform visual features on datasets with heavy occlusion \cite{vpn} and offer information that complements visual cues, our idea is to learn fine-grained geometric features (i.e., keypoints) to facilitate interactiveness prediction of human-object pairs and to enhance interaction query representation. In detail, GeoHOI improves the Transformer-based framework of STIP \cite{stip} by introducing three novel components. First, a keypoints detection module unifies the keypoint detection across different object categories, including humans, and is integrated into GeoHOI for end-to-end HOI detection. It simplifies the appearance distribution of different object classes by reconstructing object segmentation masks instead of their RGB images, allowing the network to focus on learning different shapes and enabling it to learn keypoints for arbitrary objects. As a result, it generates consistent and robust keypoint representation across different object categories. Second, a keypoint-aware interactiveness prediction module employs a graph convolutional network, capturing the holistic cues (i.e., cross-instance features) between humans and objects that complement pairwise features to effectively predict the interactiveness of human-object pairs. Third, a part attention module intends to identify informative local cues since specific interaction types are defined with detailed local information of human and object parts. This enhances the representation of interaction queries in the HOI Transformer for effectively classifying specific interactions. Thus, we exploit a self-attention mechanism to produce part-level attention, with keypoint positions serving as positional encodings. This allows the HOI classifier to focus on specific local regions that are informative to each interaction type.

We evaluate our model on two HOI benchmarks V-COCO \cite{VSRL} and HICO-DET \cite{HICO}. The proposed GeoHOI achieves superior results on both datasets. Source codes are available at \href{https://github.com/zhumanli/GeoHOI}{https://github.com/zhumanli/GeoHOI}. Our contributions are:
\begin{itemize}
    \item We introduce GeoHOI, a geometric features enhanced human-object interaction detection approach, facilitating pattern detection and measurement in images captured by vision instruments.
    \item We present a self-supervised keypoints learning method (UniPointNet) to detect keypoints for different object categories including humans in a unified manner. To the best of our knowledge, this is the first attempt that unifies keypoints detection across different object classes in HOI.
    \item We design a keypoint-aware interactiveness prediction module that incorporates holistic relationships between humans and objects. The geometric keypoint features are exploited to measure the likelihood of human-object interactions, boosting the interactiveness prediction of human-object pairs.
    \item We propose a part attention module that refines interaction query representation using self-attention, enhancing specific interaction prediction by identifying informative human and object parts.
    \item We demonstrate the effectiveness of our proposed GeoHOI by conducting experiments in public HOI detection benchmark datasets, outperforming state-of-the-art methods by a large margin of 3.4 mAP on V-COCO and 3.76 mAP on HICO-DET. We further conduct a real-world application case of post-disaster with UAVs, and GeoHOI outperforms all the baselines in terms of AP and recall.
\end{itemize}

\section{Related Work}

\subsection{Two-stage Methods}

\subsubsection{Multi-stream Approaches}
Early HOI detection models are typically implemented with a two-stage framework. In the first stage, an object detector such as Fast R-CNN \cite{FastRCNN} is used to localize instances. In the second stage, a classifier is trained to predict human-object interactions. Two-stage methods use pre-trained object detectors to simplify HOI detection, achieving a good trade-off between performance and complexity \cite{SGCN4HOI}. Earlier works focus on designing multi-branch HOI classifiers with convolutional neural networks modelling human and object appearance features and spatial layout. Gkioxari et al. \cite{InteractNet} extended Fast R-CNN by introducing a human-centric branch to predict interactions at each target object location. Chao et al. \cite{HICO} proposed a three-branch framework to model pairwise human-object appearance features and their spatial relations. Hou et al. \cite{hou2021} presented a five-branch framework with a novel fabricated compositional branch targeting the issue of long-tailed distributions of HOI interactions. These methods mainly focus on exploring the pairwise human and object features, overlooking the holistic features that could complement the pairwise ones.

Some works have exploited graph convolutional networks (GCNs) to model the relationships between humans and objects from a global perspective. Qi et al. \cite{gpnn} proposed a fully connected graph with humans and objects as nodes, and the adjacency matrix was inferred by their proposed link function. Ulutan et al. \cite{VSGNet} introduced a visual-spatial-graph network to model structural connections between instances. Similar to Qi et al., they model humans and objects as nodes. Instead of a fully connected graph, they only build connections between inter-class instances, omitting unnecessary human-human and object-object pairs. Their adjacency matrix is predicted by the visual branch. Zhang et al. \cite{zhang2021spatially} presented a spatially conditioned graph with a multi-branch fusion module computing the adjacency structure and refining graph features. GCN-based HOI methods have shown that the modelling of intra-level and inter-level HOI representations can significantly improve HOI detection performance \cite{ERNet}. The reason is that GCNs not only capture pairwise features but also infer holistic cross-instance cues, which are useful for HOI reasoning. We leverage its advantage by fusing both pairwise features and cross-instance cues to enhance HOI prediction.

\subsubsection{Geometric Features Informed Approaches}
Geometric features such as human pose and object structure provide fine-grained spatial information and have been proved to be effective in improving HOI detection performance in two-stage methods. Fang et al. \cite{PD-Net} and Wan et al. \cite{PMFNet} explored the semantic cues of human body parts with an attention module that effectively identifies the most informative body parts for HOI recognition. Wu et al. \cite{cross-person-part} proposed to extract cross-person cues for body parts, which afford useful and supplementary information for the discovery of interactiveness. Park et al. \cite{viplo} designed a graph with a pose-conditioned self-loop structure, allowing the human node embedding to be updated based on the local features of human joints. As discussed, the human pose has been well-studied in HOI detection, while the geometric features of objects such as keypoint positions are less explored. To overcome this, Zheng et al. \cite{SIGN} proposed to model the interactions between human joints and object keypoints using a graph network for capturing fine-grained spatial relationships in HOI detection. Nevertheless, the representation of object keypoints in their work (i.e., two corner points of the object bounding box) is too simple to capture object shapes or structures as it considers only the rectangular spatial scope of an object.

Efforts have been made to improve the representation of object structure in HOI detection. Zhu et al. \cite{SGCN4HOI} proposed a deterministic method for representing object keypoints which encapsulate the underlying structure of an object. They extracted an object skeleton from its segmentation mask using a morphological skeletonization algorithm and obtained its keypoints by applying the K-means clustering to the set of key points on the skeleton. This kind of non-probabilistic method is less robust in handling various object shapes, particularly when dealing with non-articulated objects, making it difficult to accurately detect keypoints across different objects. Bar et al. \cite{objectkeypoints-hoi} exploited transfer learning to estimate animal keypoints with a pre-trained object keypoints detector and adopted the interest point detection in geometry with bin girding to obtain keypoints for artefacts such as beds and computers. Ito \cite{hokem} proposed a human and object keypoint-based extension module to improve conventional HOI detection models such as \cite{VSGNet}. However, the different representations of human and object keypoints presented in these frameworks are less consistent and difficult to maintain across different objects, making them less applicable to real-world applications. In this work, we explore a self-supervised framework for learning keypoints of both humans and objects, which is a more versatile and robust approach for keypoint estimation.

\subsection{End-to-end Transformer-based Methods}
Transformers have shown superior performance in many fields including HOI detection due to their advanced network architecture and high capacity. They are first adopted in \cite{HOTR,HOITransformer,qpic} by utilizing the vanilla Transformer architecture \cite{detr} to map the parametric interaction queries into a set of HOI predictions with a bipartite matching loss. Later, Kim et al. \cite{mstr} introduced a multi-scale Transformer architecture to boost HOI detection. Recently, a multiplex relation network that disentangled Transformer decoders to encourage rich context exchange was proposed in \cite{muren}. Unlike two-stage methods that optimize instance detection and interaction detection in separate stages, these end-to-end frameworks infer human-object relationships from a global contextual perspective. They predict all elements of HOI triplets directly, significantly surpassing the performance of existing two-stage approaches. Nevertheless, the rich prior knowledge, such as spatial features \cite{VSGNet}, are not exploited in the above Transformer-based attempts.

Some studies have attempted to inject prior knowledge into Transformer architectures, to address the aforementioned limitation. Iftekhar et al. \cite{SSRT} proposed to utilize the semantic features (i.e., text embeddings) and the spatial features (i.e., the relative spatial configuration of human and object bounding box locations) to enhance the query representations of decoders. Zhang et al. \cite{stip} exploited the inter-interaction semantic structure and intra-interaction spatial structure over interaction proposals (i.e., human-object pairs) to strengthen HOI predictions. Xie et al. \cite{cql} proposed a novel category query learning approach where interaction queries are explicitly associated with specific and fixed image categories, facilitating HOI detection. We observe that spatial features, such as instance bounding boxes and human-object layout used in these works are too coarse to capture fine-grained relationships between human body parts and object parts, which have been demonstrated to be beneficial in existing two-stage HOI models \cite{SGCN4HOI,objectkeypoints-hoi,hokem}. In this paper, we leverage the geometric keypoint features to facilitate HOI classification in an end-to-end Transformer-based framework.

\begin{figure}[htbp]
\setlength{\abovecaptionskip}{1pt} 
\centering
\includegraphics[width=\linewidth]{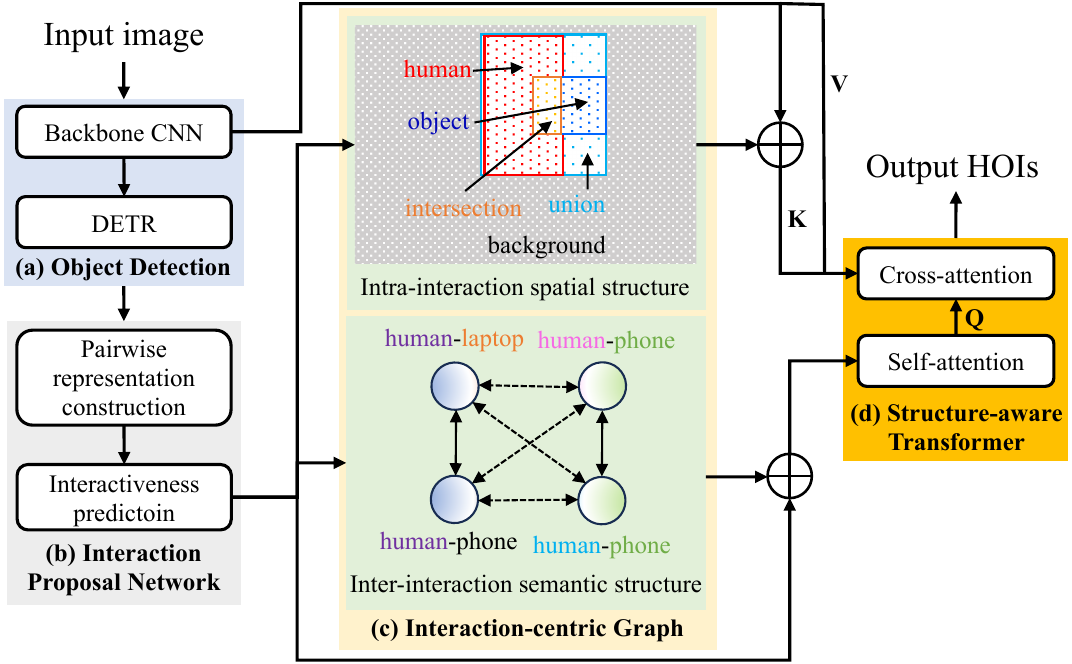}
\caption{\footnotesize{Simple illustration of STIP. The solid bi-directional arrow means whether or not two HOI triplets share the same human or object, and the dashed bi-directional arrow denotes they do not share anything. (a) Given an input image, DETR is used to detect humans and objects. (b) By constructing all possible human-object pairs, the interaction proposal network uses pairwise features to filter non-interactive ones. (c) Next, an interaction-centric graph is built to inject rich inter-interaction semantic structure and intra-interaction spatial structure. (d) Finally, a structure-aware Transformer is utilized to output a set of HOI predictions.}}
\label{Fig: stip}
\end{figure}

\begin{figure*}[htbp]
\setlength{\abovecaptionskip}{5pt} 
\centering
\includegraphics[width=0.87\textwidth]{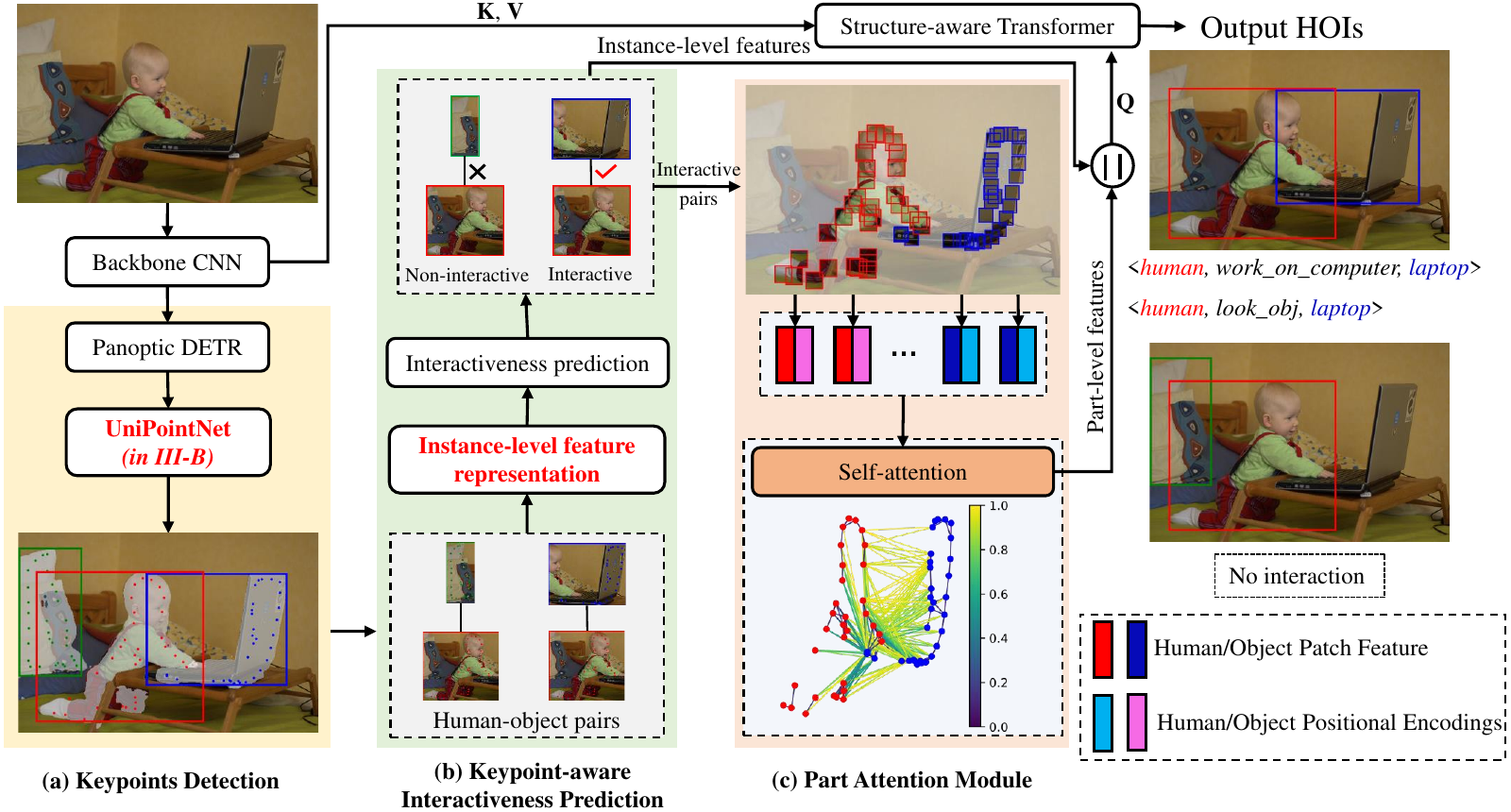}
\caption{\footnotesize{An overview of our GeoHOI framework, in which \vertcircle denotes the concatenation operation. (a) Given an image, we adopt the off-the-shelf Panoptic DETR \cite{detr} to detect the human and object instances within this image, generating their bounding boxes and segmentation masks. Based on the masks, we use our proposed UniPointNet to detect keypoints for all instances. (b) With the detected instances, the keypoint-aware interactiveness prediction module enumerates all possible human-object pairs. It then generates interactive ones with the highest interactiveness scores using coarse instance-level features, including pairwise and holistic graph features. (c) By taking all the interactive human-object pairs, we enhance their representations with human and object local patches, which are attended by self-attention. This encourages each interaction query to focus on informative human and object parts. The final concatenated representations serve as interaction queries which are then fed into the structure-aware Transformer \cite{stip} to output a set of HOI predictions.}}
\label{Fig: framework}
\end{figure*}

\section{Overview of GeoHOI}
\label{sec:Approach}
This work aims to improve end-to-end Transformer-based HOI detection networks with fine-grained geometric features of humans and objects. To this end, we propose GeoHOI. It utilizes learnable fine-grained geometric features (i.e., keypoint positions) to facilitate the interactiveness prediction of human-object pairs and to enhance interaction query representations. Inspired by the process of HOI detection with prior knowledge, we improve the structure-aware Transformer over interaction proposals (STIP \cite{stip}) by using keypoint features. As shown in Fig. \ref{Fig: stip}, STIP is an improved network over the vanilla Transformer with prior knowledge of inter-interaction (i.e., whether or not two HOI triplets share the same human or object) and intra-interaction (i.e., the layout of human and object) structure. It becomes a natural backbone of GeoHOI due to its decompose-style design of HOI predictions, i.e., interaction proposals are first generated, followed by interaction classification. Such design allows us to explore rich geometric features for effective interaction proposal generation and non-parametric interaction query representation.

Specifically, our framework introduces three novel components to STIP, i.e., keypoints detection with our novel UniPointNet, keypoint-aware interactiveness prediction module for predicting interactive human-object pairs, and part attention module to enhance interaction query representation with informative human and object local parts. We start with GeoHOI architecture (Section \ref{sec:GeoHOI}), then introduce the keypoint-aware interactiveness prediction module (\ref{sec:KIP}) and the part attention module (\ref{sec:PAM}). Section \ref{sec:UniPointNet} details the keypoints detection network (UniPointNet).

\subsection{Architecture of GeoHOI}
\label{sec:GeoHOI}
An overview of GeoHOI is shown in Fig. \ref{Fig: framework}. Given an input image $\boldsymbol{x} \in \mathbb{R}^{H \times W \times C}$, where $H$, $W$, and $C$ represent the image height, width and channels, accordingly, GeoHOI first extracts the image feature map $\boldsymbol{F_x} \in \mathbb{R}^{H^\prime \times W^\prime \times d}$ with a CNN backbone of ResNet. $\boldsymbol{F_x}$ is then sent to Panoptic DETR \cite{detr} to obtain instance detections including bounding boxes and segmentation masks. Next, the segmentation masks are fed into UniPointNet to obtain keypoints for each instance. After that, the keypoint-aware interactiveness prediction module constructs pairwise and holistic graph features for instance-level feature presentation. It then predicts and outputs interactive pairs, enhanced by local cues from keypoints in the part attention module. Finally, the structure-aware Transformer generates HOI predictions. Details are introduced in the following sections.

\subsection{Keypoint-aware Interactiveness Prediction Module}
\label{sec:KIP}
The Keypoint-aware Interactiveness Prediction (KIP) module aims to suppress non-interactive human-object pairs using coarse instance-level features. It transforms the random parametric interaction queries in the vanilla Transformer to non-parametric interaction proposals equipped with prior knowledge (e.g., instance visual features and their spatial layout), facilitating relational reasoning among interactions in HOI set prediction \cite{stip}. When learning the interactiveness of a human-object pair, visual cues can be explored not only from the targeted human and object but also from other humans and objects in the scene \cite{cross-person-part}, providing a more comprehensive understanding of the scene. However, previous works such as \cite{tin,stip} only consider target pairwise features, failing to effectively extract interactive pairs. As a potential solution, mining cues from a global cross-instance perspective, i.e., using other humans and objects as a reference, would offer helpful and supplementary information for interactiveness inference. Therefore, in addition to pairwise features, we incorporate graph features using keypoint positions measuring the geometric distance with a graph convolutional network from a global perspective, extracting cross-instance cues. We first enumerate all human-object pairs using the detected instances by Panoptic DETR, and the KIP then estimates the likelihood of interaction for each pair based on both pairwise features and holistic graph features through a multi-layer perceptron (MLP). Finally, the KIP module outputs the top-K human-object pairs with the highest probability scores. 

\begin{figure}[htbp]
	\setlength{\abovecaptionskip}{5pt} 
	\centering
	\includegraphics[scale=0.55]{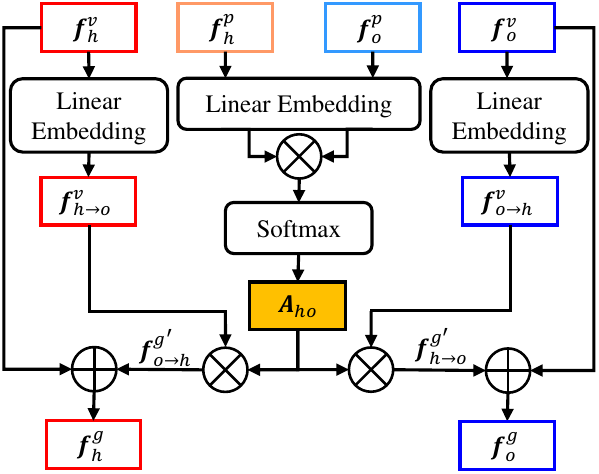}
	\caption{\footnotesize{Illustration of the graph convolution layer, in which $\otimes$ represents the tensor product, and $\oplus$ is the residual connection. The output graph features encode relationships between all humans and objects from a global perspective, with keypoint similarity measuring their connectivity.}}
	\label{Fig: gcnlayer}
\end{figure}

Concretely, for each human-object pair, the human visual feature $\boldsymbol{f}_{h}^{v}$, object visual feature $\boldsymbol{f}_{o}^{v}$, spatial feature $\boldsymbol{f}_{s}^{u}$, union feature $\boldsymbol{f}_{u}^{v}$ are represented as 256-dimensional vectors, while the object's semantic feature $\boldsymbol{f}_{o}^{c}$ (the embedding of the object class label) is a 300-dimensional vector. We refer to these as pairwise features. For the graph representation, we model humans and objects as nodes, connecting each human to all objects and each object to all humans. As shown in Fig. \ref{Fig: gcnlayer}, the node features are represented with visual features $\boldsymbol{f}_{h}^{v}$, $\boldsymbol{f}_{o}^{v}$. We use the similarities between human keypoints and object keypoints to define the adjacency matrix $\boldsymbol{A}$. This highlights that closer keypoints between a human and an object indicate a higher likelihood of interaction. Unlike the implicit adjacency matrix representation (i.e., it is predicted from instance visual features) in \cite{VSGNet}, the keypoints similarity explicitly captures the geometric distance prior knowledge between a human and an object, resulting in effective interactiveness prediction. As depicted in Fig. \ref{Fig: gcnlayer}, given the keypoint features $\boldsymbol{f}_{h}^{p} \in \mathbb{R}^{N\times2}$ of a human and $\boldsymbol{f}_{o}^{p}\in \mathbb{R}^{N\times2}$ of an object, they are first embedded by a linear layer with 128-dimensional vectors and the similarity between them is served as their edge weight $\boldsymbol{A}_{ho} \in \boldsymbol{A}$, which is expressed as follows:
\begin{equation}
    \boldsymbol{A}_{ho} = \phi(\boldsymbol{f}_{h}^{p})\otimes \phi(\boldsymbol{f}_{o}^{p}),
\end{equation}
where $\phi$ is implemented with a linear layer to encode keypoint positions, $\otimes$ denotes the dot product. Note that the edge weight between a human and an object is symmetric, i.e., $\boldsymbol{A}_{ho} = \boldsymbol{A}_{oh}$. The graph features $\boldsymbol{f}_{h}^{g}$ and $\boldsymbol{f}_{o}^{g}$ are then defined as follows:
\begin{equation}
    \boldsymbol{f}_{h}^{g} = \boldsymbol{f}_{h}^{v} + \sum\limits_{o=1}^{\hat{O}} \boldsymbol{A}_{ho} \boldsymbol{f}_{o \rightarrow h}^v,
\end{equation}
\begin{equation}
    \boldsymbol{f}_{o}^{g} = \boldsymbol{f}_{o}^{v} + \sum\limits_{h=1}^{\hat{H}} \boldsymbol{A}_{oh} \boldsymbol{f}_{h \rightarrow o}^v,
\end{equation}
where $\hat{O}$ and $\hat{H}$ are the numbers of humans and objects, $\boldsymbol{f}_{o \rightarrow h}^v$ is the projection of object visual feature $\boldsymbol{f}_{o}^{v}$ in the human space, and $\boldsymbol{f}_{h \rightarrow o}^v$ is the projection of human visual feature $\boldsymbol{f}_{h}^{v}$ in the object space.

Finally, the instance-level features for interactiveness prediction of each human-object pair in this module are obtained as the concatenation of all the features as follows:
\begin{equation}
    \boldsymbol{f}_{ho}= \boldsymbol{f}_{h}^{v} \vertcir \boldsymbol{f}_{o}^{v} \vertcir \boldsymbol{f}_{h}^{g} \vertcir \boldsymbol{f}_{o}^{g}  \vertcir \boldsymbol{f}_{u}^{s} \vertcir \boldsymbol{f}_{o}^{c} \vertcir \boldsymbol{f}_{u}^{v}.
\end{equation}

\subsection{Part Attention Module}
\label{sec:PAM}
While the instance-level features provide coarse information for interactions, specific interaction types are defined with fine-grained details. They highlight local information on human and object parts that are unlikely to be captured in instance-level features \cite{PMFNet}. In addition, the fine-grained correlations among human body parts and object parts (e.g., the spatial layout between the human hands and the laptop keyboard shown in Fig. \ref{Fig: framework}) implicitly depict the consistent spatial, scale, and co-occurrence relationships between humans and objects, providing a finer granularity context information of an image \cite{objectkeypoints-hoi}. However, existing works \cite{RPNN,cross-person-part} only consider human body parts while overlooking object structure parts. 

To address the aforementioned limitation, we introduce a Part Attention Module (PAM) designed to identify the most relevant parts of both humans and objects for detecting a specific interaction category. We use self-attention to learn the part-level features of a given human-object pair, enabling each part to aggregate information from all other parts, regardless of their distance or position. This allows the network to extract richer and more comprehensive context features, leading to a deeper understanding of the scene. This module serves to enhance the interaction query representation of each selected interactive human-object pair, improving the effectiveness of classifying particular interactions.

In detail, given the human keypoints $\boldsymbol{f}_{h}^p = \left\{ \boldsymbol{f}_{h}^{p1} \dots, \boldsymbol{f}_{h}^{pN} \right\}$, we define a local region 
$\boldsymbol{x}_{pi} \in \mathbb{R}^4$ for each keypoint $\boldsymbol{p}_{i}^h$, it is centered at $\boldsymbol{p}_{i}^h$ and has a size $\gamma$ proportional to the size of the human bounding box. We adopt RoI-Align \cite{mask-rcnn} to generate local patch features and rescale them to a resolution of $R_p \times R_p$. We apply the same operations to object keypoints $\boldsymbol{f}_{o}^p = \left\{ \boldsymbol{f}_{o}^{p1} \dots, \boldsymbol{f}_{o}^{pN} \right\}$ to generate their local patch features as well. For the sake of simplicity, we denote the extracted patch features of humans and objects as $\boldsymbol{f}^{p^\prime} = \left\{ \boldsymbol{f}^{p^{\prime}1} \dots, \boldsymbol{f}^{p^\prime {2N}} \right\}$. In addition, we embed each keypoint as positional encodings to its corresponding patch. By doing this, the model can capture more detailed spatial relationships and configurations within each human-object pair. It also ensures a richer representation of the data, allowing the model to make more context-aware predictions of specific interaction types.

We then represent patch features of each human-object pair, integrated with their corresponding positional encodings, as a sequence of queries $\hat{\boldsymbol{q}}=\left(\hat{\boldsymbol{q}}_1, \ldots, \hat{\boldsymbol{q}}_{2N}\right)$, keys $\hat{\boldsymbol{k}}=\left(\hat{\boldsymbol{k}}_1, \ldots, \hat{\boldsymbol{k}}_{2N}\right)$, and values  $\hat{\boldsymbol{v}}=\left(\hat{\boldsymbol{v}}_1, \ldots, \hat{\boldsymbol{v}}_{2N}\right)$. Following the self-attention mechanism \cite{attention}, each patch is computed by aggregating all values weighted with attention, and an attended patch feature is represented as follows:
\begin{equation}
    \boldsymbol{f}^{\hat{p}}_i=\sum_j \alpha_{ij}\left(\boldsymbol{W}_{\hat{v}} \hat {\boldsymbol{v}}_j\right), 
\end{equation}
where each $\alpha_{i j}=\frac{\exp \left(e_{ij}\right)}{\sum_j \exp \left(e_{ij}\right)}$ is the normalized attention weight with softmax. Here the primary attention weight $e_{ij}$ is the scaled dot-product between each key $\hat{\boldsymbol{k}}$ and query $\hat{\boldsymbol{q}}$:
\begin{equation}
    e_{ij}=\frac{\left(\boldsymbol{W}_{\hat{q}} \boldsymbol{\hat{q}}_i\right)^T\left(\boldsymbol{W}_{\hat{k}} \boldsymbol{\hat{k}}_j\right)}{\sqrt{d_{key}}},
\end{equation}
note that $\boldsymbol{W}_{\hat{q}}$, $\boldsymbol{W}_{\hat{k}}$, $\boldsymbol{W}_{\hat{v}}$ are learnable embedding matrices, and $d_{key}$ is the embedding dimension of keys.

The attended local part feature for a human-object pair is calculated by concatenating all patches:
\begin{equation}
    \boldsymbol{f}^{\hat{p}} = 
 \boldsymbol{f}^{\hat{p}}_1 \vertcir \boldsymbol{f}^{\hat{p}}_2 \vertcir \dots \vertcir \boldsymbol{f}^{\hat{p}}_{2N}.
\end{equation}

Finally, each interaction query $\boldsymbol{q} \in Q$ is represented by the fusion of instance-level interactiveness features and the attended part features:
\begin{equation}
  \boldsymbol{q} = \boldsymbol{f}_{ho} \vertcir \boldsymbol{f}^{\hat{p}}.
\end{equation}

They are fed into the structure-aware Transformer \cite{stip} for HOI classification.

\subsection{Training and Inference}
We follow the training and inference procedure of the STIP \cite{stip}. The KIP module is optimized with focal loss (FL) \cite{focalloss}:
\begin{equation}
    L_{interactiveness}=\frac{1}{\sum_{i=1}^{\hat{N}} z_i} \sum_{i=1}^{\hat{N}} F L\left(\hat{z}_i, z_i\right),
\end{equation}
where $\hat{N}$ is the number of sampled human-object pairs, $z_i \in \{0, 1\}$ denotes the existence of ground-truth interaction, and $\hat{z}_i$ is the predicted interactiveness score. For each of the output human-object pair of KIP, the focal loss is also used as the multi-label classification loss to train the possible interactions:
\begin{equation}
    L_{class}=\frac{1}{\sum_{i=1}^{\hat{N}} \sum_{j=1}^{\hat{C}} y_{i j}} \sum_{i=1}^{\hat{N}} \sum_{j=1}^{\hat{C}} F L\left(\hat{y}_{i j}, y_{i j}\right),
\end{equation}
where $\hat{C}$ is the number of interaction classes, $y_{ij} \in \{0, 1\}$ indicates the ground-truth interaction class, and $\hat{y}_{i j}$ is the predicted probability of $j$-th interaction class. The overall training objective of our GeoHOI integrates the above interactiveness loss and the interaction classification loss:
\begin{equation}
    L_{GeoHOI}=L_{interactiveness}+L_{class}.
\end{equation}

\begin{figure*}[htbp]
\setlength{\abovecaptionskip}{5pt} 
\centering
\includegraphics[width=0.8\textwidth]{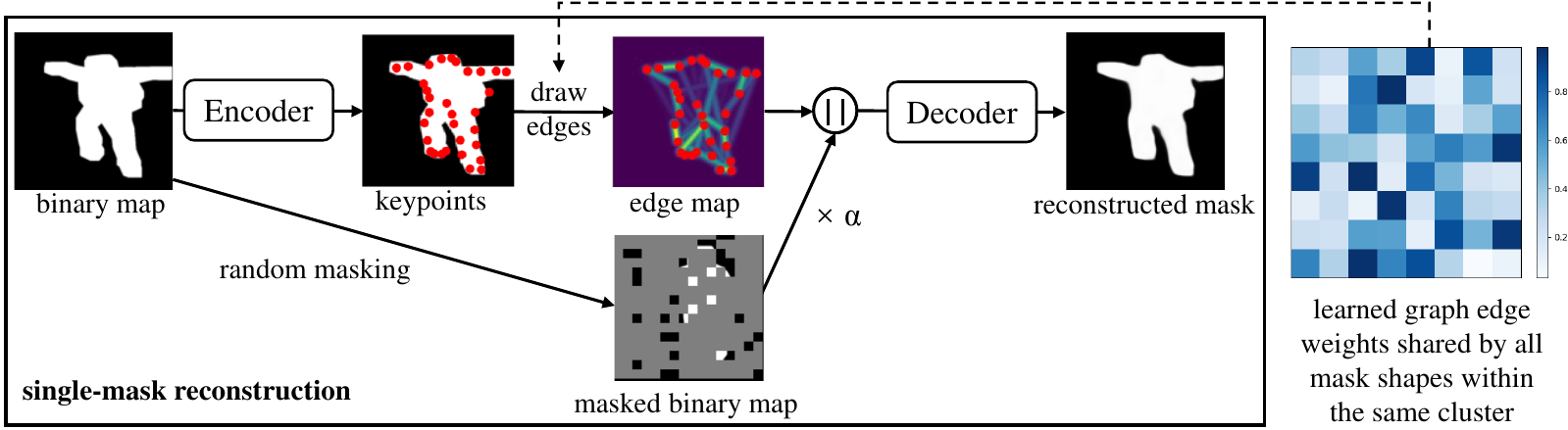}
\caption{\footnotesize{Overview of the self-supervised keypoints learning framework (UniPointNet). Given an object segmentation, we detect the keypoints with learnable graph edge weights by reconstructing its binary mask. The edge weights are represented by a color matrix and are shared across segmentation masks within clusters of similar shapes and structures. The masked segmentation binary map provides minimal appearance information, forcing the network to focus on learning keypoints that are important for representing the structure and shape of an object.}}
\label{Fig: framework_recon}
\end{figure*}

\section{self-supervised object keypoint detection}
\label{sec:UniPointNet}
As a core of our model, we propose leveraging keypoints as fine-grained geometric features of both humans and objects, to facilitate HOI prediction, but it is essential to detect these keypoints before utilizing them. Existing keypoints detection models typically focus on a single object class (e.g., human pose estimation \cite{humanpose}) 
rather than common objects. The difficulties lie in the complexity of distinct spatial structures and appearance distributions exhibited by various objects and the limited annotation availability. In addition, there is very limited work in detecting keypoints across different objects in HOI \cite{SGCN4HOI} due to a large number of object categories (e.g., 80 common object categories in MS-COCO \cite{COCO}) and occlusions. 

To address this challenge, we propose UniPointNet which can detect keypoints for arbitrary objects. We employ the self-supervised keypoints learning framework of AutoLink \cite{autolink}. While AutoLink was proposed to learn keypoints for single object classes, our goal is to detect keypoints across all classes present in the HOI task. To this end, we make two key changes to AutoLink. First, we feed object segmentation masks into the network instead of RGB images. This eliminates the appearance variations across different object classes, simplifying their appearance distribution. As a result, the network can focus on learning object shapes and structures. Second, instead of using an individual edge graph with shared graph weight to align all samples, we opt for a set of edge graphs with different graph weights, aligning samples within their respective clusters. This design accommodates object masks with significant variations, thus allowing the network to detect keypoints across a diverse range of object categories.

Using such a network to detect keypoints for humans and all the other object classes is advantageous. First, it unifies keypoints detection for different object classes within a single network, which is more applicable in real-world applications in which diverse object types are often involved. Second, it ensures the consistency of keypoints distribution across different object categories including humans, resulting in a unified and consistent keypoints representation that facilitates network learning. Third, unlike the common keypoints representation in occluded cases (e.g., zeros for occluded or invisible joints of a human), all the detected keypoints in our UniPointNet contribute to the representation of an object's shape. This guarantees a more robust keypoints representation when objects are partially visible.

\subsection{Architecture of UniPointNet}
An overview of UniPointNet is shown in Fig. \ref{Fig: framework_recon}. Given an object segmentation binary map $\boldsymbol{B} \in \mathbb{R}^{H \times W \times 1}$ with a height of $H$ and a width of $W$, our goal is to learn a set of keypoints $\kappa = \left\{ \boldsymbol{k}_i|i=1, 2, 3, \dots, N; \boldsymbol{k}_i \in[0,1] \times[0,1] \subset \mathbb{R}^2\right\}$, where $N$ is the number of keypoints. As per \cite{autolink}, keypoints are detected by the encoder with ResNet and upsampling, and each pair of keypoints is connected with a differentiable edge \cite{autolink73}. This kind of graph connectivity defines a unique structure of a group of objects with similar shapes that share the same cluster label, learned in a self-supervised manner. The edge map $\boldsymbol{E} \in \mathbb{R}^{H \times W}$ is concatenated with the masked binary map $\boldsymbol{B}_m \in \mathbb{R}^{H \times W \times 1}$ along the channel dimension, and fed into the decoder to obtain the reconstructed segmentation binary map $\boldsymbol{B^\prime}$. Detailed encoder and decoder network architectures can be referred to \cite{autolink}.

\subsection{Segmentation Structure Representation}
Here, we introduce keypoints representation and the edge map generation. $\mathcal{H} = \left\{ \boldsymbol{h}_i|i=1, 2, 3, \dots, N; \boldsymbol{h}_i \in\mathbb{R}^{H \times W}\right\}$ is the $N$ heatmaps generated by the Encoder from the input mask. The keypoint $\boldsymbol{k}_i$ is obtained by the differentiable soft-argmax function, 
\begin{equation}
    \boldsymbol{k}_i = \sum\limits_{\boldsymbol{p}} \psi(\boldsymbol{h}_i)\boldsymbol{p},
\end{equation} 
where $\psi(\boldsymbol{h}_i)$ is the $Softmax$ operation on a single heatmap $\boldsymbol{h}_i$, defined as,
\begin{equation}
    \psi(\boldsymbol{h}_i) = \frac{exp(\boldsymbol{h}_{i}(\boldsymbol{p}))}{ \sum\limits_{j=1}^N exp(\boldsymbol{h}_{j}(\boldsymbol{p}))},
\end{equation}
where $\boldsymbol{p}$ is normalized pixel coordinates.

According to \cite{autolink}, a differentiable edge map $\boldsymbol{E}_{ij}$ is generated for any two keypoints $\boldsymbol{k}_i$ and $\boldsymbol{k}_j$, by assigning a value of 1 to pixels on the edge connecting the keypoints. For other pixels, their values decrease exponentially based on their distance to the line. The edge map $\boldsymbol{E}_{ij}$ is a Gaussian that extends along the line \cite{autolink73}, and it is formally expressed as, 
\begin{equation}
    \boldsymbol{E}_{i j}(\boldsymbol{p})=\exp \left(d_{i j}^2(\boldsymbol{p}) / \sigma^2\right),
\end{equation}
where the hyperparameter $\sigma$ controls the thickness of the line, and $d_{i j}^2(\boldsymbol{p})$ is the $L_2$ distance between the pixel $\boldsymbol{p}$ and the line from keypoints $\boldsymbol{k}_i$ and $\boldsymbol{k}_j$. According to the location of the pixel $\boldsymbol{p}$, i.e., before the starting keypoint $\boldsymbol{k}_i$, between the starting keypoint $\boldsymbol{k}_i$ and the ending keypoint $\boldsymbol{k}_j$, or after the ending keypoint $\boldsymbol{k}_j$, it is defined as, 
\begin{equation}
\begin{split}
    & \boldsymbol{d}_{i j}(\boldsymbol{p})=\left\{\begin{array}{lr}\left\|\boldsymbol{p}-\boldsymbol{k}_i\right\|_2 & \text { if } t \leq 0, \\ \left\|\boldsymbol{p}-\left(\boldsymbol{k}_i+t \boldsymbol{k}_j\right)\right\|_2 & \text { if } 0<t<1, \\ \left\|\boldsymbol{p}-\boldsymbol{k}_j\right\|_2 & \text { if } t \geq 1,\end{array} \quad\right. 
    \\ & where \quad t=\frac{\left(\boldsymbol{p}-\boldsymbol{k}_i\right) \cdot\left(\boldsymbol{k}_j-\boldsymbol{k}_i\right)}{\left\|\boldsymbol{k}_i-\boldsymbol{k}_j\right\|_2^2}.
\end{split}
\end{equation}

The final edge map $\boldsymbol{E} \in \mathbb{R}^{H \times W}$ is obtained by taking the maximum at each pixel of all the heatmaps,
\begin{equation}
    \boldsymbol{E}(\boldsymbol{p})=\max _{i j} w_{i j} \boldsymbol{E}_{i j}(\boldsymbol{p}),
\end{equation}
where $w_{i j}$ is a learnable edge weight. 
As explained in \cite{autolink}, opting for the maximum value at each pixel helps untangle the edge weights from the convolution kernel weights and generates better performance.

\subsection{Segmentation Reconstruction}
The masked segmentation $\boldsymbol{B}_m$ is obtained by randomly masking out $90\%$ of the input segmentation. It is then concatenated with the edge map and is fed into the decoder to reconstruct the original segmentation,
\begin{equation}
    \boldsymbol{B^\prime} = Decoder(\alpha\boldsymbol{B}_m \vertcir \boldsymbol{E}),
\end{equation}
where $\vertcir$ denotes concatenation along the channel dimension and the parameter $\alpha$ is a learnable factor that adjusts for the variation in edge weight magnitude during training and is initialized to 1. The $L_1$ loss and VGG perceptual loss \cite{percloss} are used to minimize the difference between the original segmentation and the reconstructed one,
\begin{equation}
    \mathcal{L}=\frac{1}{M} \sum_{i=1}^M (|B_i-B_i^{\prime}| + \left\|\Gamma\left(B_i\right)-\Gamma\left(B_i^{\prime}\right)\right\|_2^2),
\end{equation}
where $M$ represents the total number of examples, and $\Gamma$ indicates the feature extractor, i.e., the VGG network.

\section{Experiments} 
In this section, we introduce HOI benchmark datasets V-COCO \cite{VSRL} and HICO-DET \cite{HICO}, followed by experimental settings and implementation details. We then evaluate our proposed model against state-of-the-art approaches and provide insights on per-class performance by comparing it with the backbone STIP. Finally, we present ablation studies on the selection of the number of keypoints and the impact of individual component designs of our model.

\subsection{Datasets}
\textbf{V-COCO} is a popular HOI detection dataset and is a subset of MS-COCO \cite{COCO} including 29 different action classes. It consists of 10,346 images, with 2533 images for training, 2867 images for validating, and 4946 images for testing. Following the settings in previous works \cite{HOTR,stip,SGCN4HOI}, we apply the Average Precision ($AP_{role}$) metric over 24 interactions for the evaluation. Five actions are omitted as one of them has limited samples and the other four have no object associated with humans. Two types of $AP_{role}$ (i.e., $AP_{role}^{\#1}$ and $AP_{role}^{\#2}$) are reported under different scenarios with different scoring criteria for cases where objects are occluded. Concretely, in the scenario of $AP_{role}^{\#1}$, the occluded object bounding box must be predicted as empty, i.e., $[0,0,0,0]$. In contrast, in scenario $AP_{role}^{\#2}$, the occluded object is ignored. A human-object pair is considered a true positive if the predicted bounding boxes for both the human and the object have an Interaction-over-Union (IoU) ratio greater than 0.5 with their corresponding ground-truth annotation and the interaction category is accurate.

\textbf{HICO-DET} is a larger HOI detection dataset consisting of 47,051 images with 37,535 training and 9,515 testing images. It has 600 annotated human-object interactions and covers the same 80 object categories in MS-COCO \cite{COCO}. We follow previous works \cite{HOITransformer,stip} and report in two different settings, i.e., \textit{Default} and \textit{Known Object}. The \textit{Default} setting represents the evaluation of AP across all testing images, whereas the \textit{Known Object} setting calculates the AP of each object solely on images that contain that object class. We report the AP for each setting over three different sets of HOI categories based on the number of training samples, i.e., \textbf{Full} (all 600 HOI categories), \textbf{Rare} (138 HOI categories that have less than 10 training samples), and \textbf{Non-Rare} (462 HOI categories with at least 10 training samples).

\begin{table*}[htbp]
\scriptsize
\centering
\caption{Performance comparison with end-to-end methods of mAP on V-COCO and HICO-DET. The best results are marked in \textbf{bold} and the second best results are marked with \underline{underline}.}
\label{table:performance_end}
\begin{tabular}{c c c c c c c c c c c} 
\hline 
\multirow{3}{*}{Method} & \multirow{3}{*}{Published In} & \multirow{3}{*}{Backbone} & \multicolumn{2}{c}{\textbf{V-COCO}} & \multicolumn{6}{c}{\textbf{HICO-DET}} \\
\cline{4-11} 
& & & \multirow{2}{*}{${AP}_{role}^{\#1}$} & \multirow{2}{*}{${AP}_{role}^{\#2}$} & \multicolumn{3}{c}{Default} & \multicolumn{3}{c}{Known Object} \\
\cline{6-11} 
& & & & & Full & Rare & Non-Rare & Full & Rare & Non-Rare \\
\hline

UnionDet \cite{uniondet} & ECCV 2020 & R50-FPN & 47.5 & 56.2 & 17.58 & 11.72 & 19.33 & 19.76 & 14.68 & 21.27 \\
IPNet \cite{IPNet} & CVPR 2020 & HG-104 & 51.0 & - & 19.56 & 12.79 & 21.58 & 22.05 & 15.77 & 23.92 \\
GGNet \cite{ggnet} &  CVPR 2021 & HG-104 & 54.7 & - & 29.17 & 22.13 & 30.84 & 33.50 & 26.67 & 34.89 \\
HOTR \cite{HOTR} & CVPR 2021 & R50 & 55.2 & 64.4 & 23.46 & 16.21 & 25.60 & - & - & - \\
QPIC \cite{qpic} & CVPR 2021 & R50 & 58.8 & 61.0 & 29.07 & 21.85 & 31.23 & 31.68 & 24.14 & 33.93 \\
DSSF \cite{dssf} & IEEE TIM 2022 & HG-104 & 57.6 & - & 25.23 & 18.72 & 27.17 & 28.53 & 21.68 & 30.57 \\
MSTR \cite{mstr} & CVPR 2022 & R50 & 62.0 & 65.2 & 31.17 & 25.31 & 32.92 & 34.02 & 28.83 & 35.57 \\
ERNet \cite{ERNet} & IEEE TIP & EfficientNet &64.2 & - & 31.57 & 26.76 & 33.10 & - & - & - \\

MUREN \cite{muren} & CVPR 2023 & R50 & \underline{68.8} & 71.0 & \underline{32.87} & 28.67 & \underline{34.12} & \underline{35.52} & 30.88 & \underline{36.91} \\

STIP \cite{stip} & CVPR 2022 & R50 & 66.0 & 70.7 & 28.81 & 27.55 & 29.18 & 32.28 & 31.07 & 32.64 \\ 

STIP* \cite{stip} & CVPR 2022 & R50 & - & - & 32.22 & 28.15 & 33.43 & 35.29 & 31.43 & 36.45 \\ \hline

GeoHOI (Ours) & - & R50 & 67.8 & \underline{73.3} & 30.07 & \underline{29.72} & 30.13 & 33.36 & \underline{32.97} & 33.43 \\ 

GeoHOI* (Ours) & - & R50 & \textbf{69.4} & \textbf{74.4} & \textbf{35.05} & \textbf{33.01} & \textbf{35.71} & \textbf{37.12} & \textbf{34.79} & \textbf{37.97} \\ \hline
\end{tabular}
\end{table*}

\begin{table*}[htbp]
\scriptsize
\centering
\caption{Performance comparison with two-stage methods of mAP on V-COCO and HICO-DET. The notation is the same as in Table \ref{table:performance_end}.}
\label{table:performance_twostage}
\begin{tabular}{c c c c c c c c c c c} 
\hline 
\multirow{3}{*}{Method} & \multirow{3}{*}{Published In} & \multirow{3}{*}{Backbone} & \multicolumn{2}{c}{\textbf{V-COCO}} & \multicolumn{6}{c}{\textbf{HICO-DET}} \\
\cline{4-11} 
& & & \multirow{2}{*}{${AP}_{role}^{\#1}$} & \multirow{2}{*}{${AP}_{role}^{\#2}$} & \multicolumn{3}{c}{Default} & \multicolumn{3}{c}{Known Object} \\
\cline{6-11} 
& & & & & Full & Rare & Non-Rare & Full & Rare & Non-Rare \\
\hline 

InteractNet \cite{InteractNet} & CVPR 2018 & R50-FPN & 40.0 & 48.0 & 9.94 & 7.16 & 10.77 & - & - & - \\
TIN \cite{tin} & CVPR 2019 & R50 & 48.7 & - & 17.22 & 13.51 & 18.32 & 19.38 & 15.38 & 20.57 \\
DRG \cite{drg} & ECCV 2019 & R50-FPN & 51.0 & - & 24.53 & 19.47 & 26.04 & 27.98 & 23.11 & 29.43 \\
FCMNet \cite{fcmnet} & ECCV 2020 & R50 & 53.1 & - & 20.41 & 17.34 & 21.56 & 22.04 & 18.97 & 23.12 \\
IDN \cite{idn} & NeurIPS 2020 & R50 & 53.3 & 60.3 & 23.36 & 22.47 & 23.63 & 26.43 & 25.01 & 26.85 \\

iHOI \cite{iHOI}  & IEEE TMM 2020 & R50-FPN & 45.8 & - & 13.39 & 9.51 & 14.55 & - & - & - \\
ACP++ \cite{ACP++} & IEEE TIP 2021 & R152 & 53.2 & - & 18.90 & 16.80 & 19.52 & 24.78 & 23.87 & 25.05 \\
HRNet \cite{HRNet} & IEEE TIP 2021 & R152 & 53.1 & - & 21.93 & 16.30 & 23.62 & 25.22 & 18.75 & 27.15 \\
IPGN \cite{IPGN} & IEEE TIP 2021 & R50-FPN & 53.8 & - & 21.26 & 18.47 & 22.07 & - & - & - \\
CP-HOI \cite{CP-HOI} & IEEE TPAMI 2022 & R50 & 50.4 & - & 19.42 & 13.98 & 20.91 & 22.01 & 15.73 & 22.80 \\

UPT \cite{upt} & CVPR 2022 & R101-DC5 & 61.3 & 67.1 & 32.62 & 28.62 & 33.81 & 36.08 & 31.41 & 37.47 \\
ViPLO \cite{viplo} & CVPR 2023 & ViT & 62.2 & 68.0 & \textbf{37.22} & \textbf{35.45} & \textbf{37.75} & \textbf{40.61} & \textbf{38.82} & \textbf{41.15} \\ \hline

GeoHOI (Ours) & - & R50 & \underline{67.8} & \underline{73.3} & 30.07 & 29.72 & 30.13 & 33.36 & 32.97 & 33.43 \\ 
GeoHOI* (Ours) & - & R50 & \textbf{69.4} & \textbf{74.4} & \underline{35.05} & \underline{33.01} & \underline{35.71} & \underline{37.12} & \underline{34.79} & \underline{37.97} \\ \hline
\end{tabular}
\end{table*}

\subsection{Implementation Details}
To train UniPointNet, we extract object segmentation masks from the COCO dataset \cite{COCO}. Masks with a ratio less than 0.2 relative to the image are discarded, since we aim to learn object shapes and structures and these tiny masks do not contain enough pixels to compute shape features. We finally collected a total of 50,238 training samples. We then apply ResNet to group all samples into 100 clusters with K-means clustering. Each cluster is associated with a unique set of graph weights during training, aligning the shape of samples within that cluster. Following AutoLink \cite{autolink}, the network is trained for 20k iterations with Adam optimizer, a learning rate of $10^{-4}$, a batch size of 64, and the edge thickness of $\sigma^2=5 e-5$. During inference, an input sample is assigned a cluster label based on its distance to each cluster centroid. Subsequently, its keypoints are detected using the corresponding graph weights. The number of keypoints can range from 4 to 48.

We adopt the object detector Panoptic DETR \cite{detr} pre-trained over MS-COCO, for both object bounding box detection and segmentation. It provides segmented inputs to UniPointNet, thereby enabling us to seamlessly incorporate UniPointNet into GeoHOI. The UniPointNet is then utilized as a pre-trained component of GeoHOI, detecting HOIs in an end-to-end manner. The backbone RestNet-50 is used for image feature extraction. We present results for two variations of our proposed method: GeoHOI and GeoHOI*. GeoHOI is trained for the HOI detector only with frozen parameters in Panoptic DETR, whereas GeoHOI* is trained with joint fine-tuning of both the object detector and HOI detector in an alternate manner. In the experiments, same as STIP, we output top-32 interactive human-object pairs of the Keypoints-aware Interactiveness Prediction module. Following the previous practice in \cite{PMFNet}, the RoI-Align in PAM is set to a resolution of $R_p = 5$, and the size of human and object patches is $\gamma=0.1$ of their respective instance bounding box height and width. Then all the patch features are scaled to $5\times5$. The whole architecture is trained for 30 epochs over a single NVIDIA A100 GPU with a mini-batch size of 6, initial learning rate $5\times10^{-5}$, and AdamW optimizer.

\begin{figure*}[htbp]
\setlength{\abovecaptionskip}{1pt} 
\centering
\includegraphics[width=\textwidth]{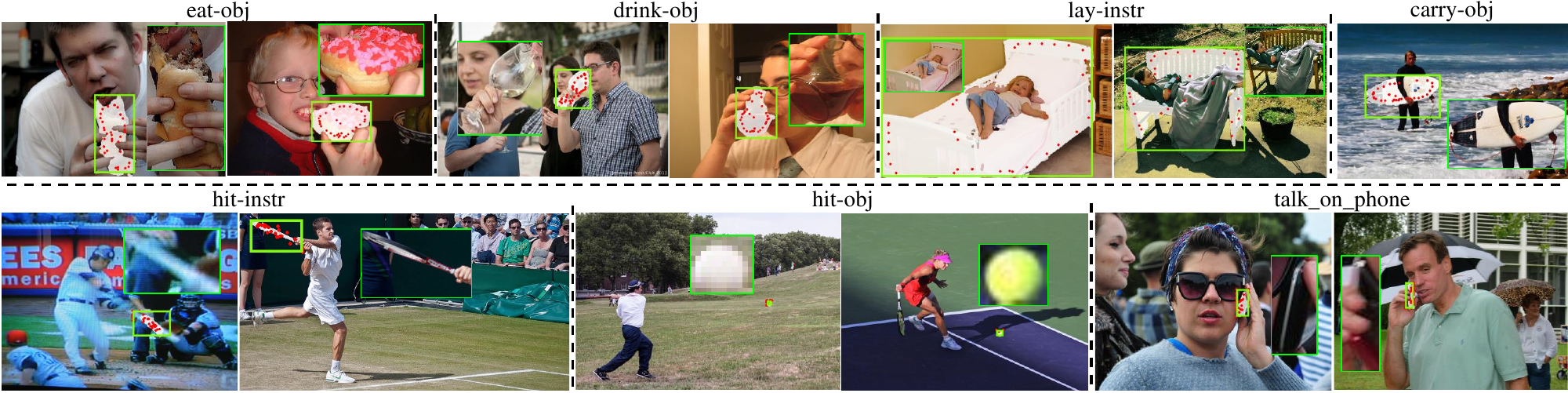}
\caption{\footnotesize{Qualitative results. The upper row showcases the effectiveness of the keypoints representation, while the lower row depicts failure cases.}}
\label{Fig: case-study}
\end{figure*}

\subsection{Comparisons with State of the Art}

We evaluate the performance of GeoHOI and compare it with state-of-the-art models, including methods that use geometric features of both humans and objects.  

Table \ref{table:performance_end} shows the performance comparison with end-to-end methods. For V-COCO, our method beats all existing end-to-end methods by a large margin in both scenarios. In particular, compared with MUREN \cite{muren}, which is the previous state-of-the-art method, GeoHOI achieves a significant performance gain of 0.6 mAP in $AP_{role}^{\#1}$and 3.4 mAP in $AP_{role}^{\#2}$. For HICO-DET, GeoHOI achieves consistent performance gains and surpasses all the previous state-of-the-art methods. These results indicate our method's effectiveness in capturing the holistic cross-instance cues between humans and objects using their keypoints through graph convolutional networks and enhancing interaction query representations with local patches. 

In table \ref{table:performance_twostage}, we compare GeoHOI against two-stage methods. Our GeoHOI outperforms all the existing two-stage methods on V-COCO. Compared with the latest method ViPLO \cite{viplo}, it obtains large performance improvements of 7.2 mAP in $AP_{role}^{\#1}$and 6.4 mAP in $AP_{role}^{\#2}$. This is mainly because most of these two-stage methods use CNNs or vanilla Transformers for HOI classification, leading to limited model capacity or prior knowledge. For HICO-DET, we also achieve comparable performance to previous state-of-the-art methods. Compared to ViPLO, the performance gain is not as noticeable as on V-COCO. Considering the complexity and end-to-end nature of GeoHOI, we follow STIP to use the lightweight ResNet for image feature extraction, while ViPLO employs the advanced Transformer backbone ViT \cite{vit} in their first stage of object detection. We believe ViT has a larger capacity than ResNet and is superior for handling the larger HICO-DET.

\begin{table}[htbp]
\scriptsize
    \centering
    \caption{Comparison of performance with object-structure-aware methods on V-COCO.}
    \label{table: object-structure-method}
    \begin{tabular}{c c c c}\hline
    Method & Published In & ${AP}_{role}^{\#1}$ & ${AP}_{role}^{\#2}$ \\ \hline \hline
    SGCN4HOI \cite{SGCN4HOI} & IEEE SMC 2022 & 53.1 & 57.9 \\ 
    Liu et al. \cite{Liu-2022} & Pattern Recognition 2022 & 52.3 & -\\
    HOKEM \cite{hokem} & IEEE ICIP 2023 & 54.6 & 59.7 \\ 
    ObjectPart \cite{objectkeypoints-hoi} & Pattern Recognition 2023 & 62.5 & - \\ \hline
    GeoHOI (Ours) & - & \textbf{67.8} & \textbf{73.3} \\ 
    \hline
		
    \end{tabular}
\end{table}

Table \ref{table: object-structure-method} compares our results with existing methods using human and object keypoints on V-COCO. We omit comparison on HICO-DET because most of these methods did not provide results on this dataset. For fairness, we compare GeoHOI without fine-tuning the object detector against these methods. The proposed GeoHOI outperforms all of them by a marked margin in both $AP_{role}^{\#1}$(5.3 mAP) and $AP_{role}^{\#2}$ (13.6 mAP). Demonstrating the effectiveness of GeoHOI, i.e., by taking advantage of both the advanced Transformer architecture and the fine-grained geometric keypoints, it boosts HOI detection. 

\begin{table}[htbp]
    \scriptsize
    \centering
    \caption{Per-class ${AP}_{role}^{\#1}$ performance comparisons with STIP.}
    \label{table: perclass}
    \begin{tabular}{ c c c c c}\hline
    HOI Class & STIP \cite{stip} & GeoHOI \\ \hline
    hold-obj (\#pos = 3608) &56.07 & \textbf{57.68} ($\uparrow 0.71$) \\ 
    sit-instr (\#pos = 1916) &56.44 & \textbf{58.99} ($\uparrow 2.55$) \\ 
    ride-instr (\#pos = 556) & 75.35 & \textbf{75.52} ($\uparrow 0.17$) \\ 
    look-obj (\#pos = 3347) &\textbf{46.09} ($\uparrow 0.88$) &45.21 \\ 
    hit-instr (\#pos = 349) &\textbf{82.21} ($\uparrow 2.91$) &79.30 \\
    hit-obj (\#pos = 349) &\textbf{78.22} ($\uparrow 3.43$) & 74.79 \\ 
    eat-obj (\#pos = 521) &65.48 & \textbf{74.52} ($\uparrow 9.04$) \\ 
    eat-instr (\#pos = 521) &77.44 & \textbf{80.65} ($\uparrow 3.21$) \\ 
    jump-instr (\#pos = 635) &\textbf{82.37} ($\uparrow 1.09$) & 81.28 \\ 
    lay-instr (\#pos = 387) &61.50 &\textbf{69.70} ($\uparrow 8.2$) \\ 
    talk\_on\_phone (\#pos = 285) &\textbf{57.30} ($\uparrow 2.75$) &54.55 \\ 
    carry-obj (\#pos = 472) &39.56 &\textbf{48.97} ($\uparrow 9.41$) \\ 
    throw-obj (\#pos = 244) &55.25 &\textbf{56.22} ($\uparrow 0.97$) \\ 
    catch-obj (\#pos = 246) &\textbf{53.59} (0.98) &52.61 \\ 
    cut-instr (\#pos = 269) &57.00 &\textbf{57.57} ($\uparrow 0.57$) \\ 
    cut-obj (\#pos = 269) &70.35 &\textbf{72.33} ($\uparrow 1.98$) \\ 
    work\_on\_comp (\#pos = 410) &75.80 &\textbf{79.03} ($\uparrow 3.23$) \\ 
    ski-instr (\#pos = 424) &\textbf{55.03} ($\uparrow 0.83$) &54.20 \\ 
    surf-instr (\#pos = 486) &79.88 &\textbf{86.51} ($\uparrow 6.63$) \\ 
    skateboard-instr (\#pos = 417) &\textbf{92.66} ($\uparrow 1.59$) &91.07 \\ 
    drink-instr (\#pos = 82) &55.30 &\textbf{65.20} ($\uparrow 9.9$) \\ 
    kick-obj (\#pos = 180) &73.89 &\textbf{76.20} ($\uparrow 2.31$) \\ 
    read-obj (\#pos = 111) &\textbf{51.92} ($\uparrow 0.89$)&51.03 \\ 
    snowboard-instr (\#pos = 277) &82.54  &\textbf{84.30} ($\uparrow 1.76$) \\ \hline
    Average &65.89 & \textbf{67.81} ($\uparrow 1.92$) \\ \hline
		
    \end{tabular}
\end{table}

\begin{figure}[htbp]
\setlength{\abovecaptionskip}{1pt} 
\centering
\includegraphics[width=\linewidth]{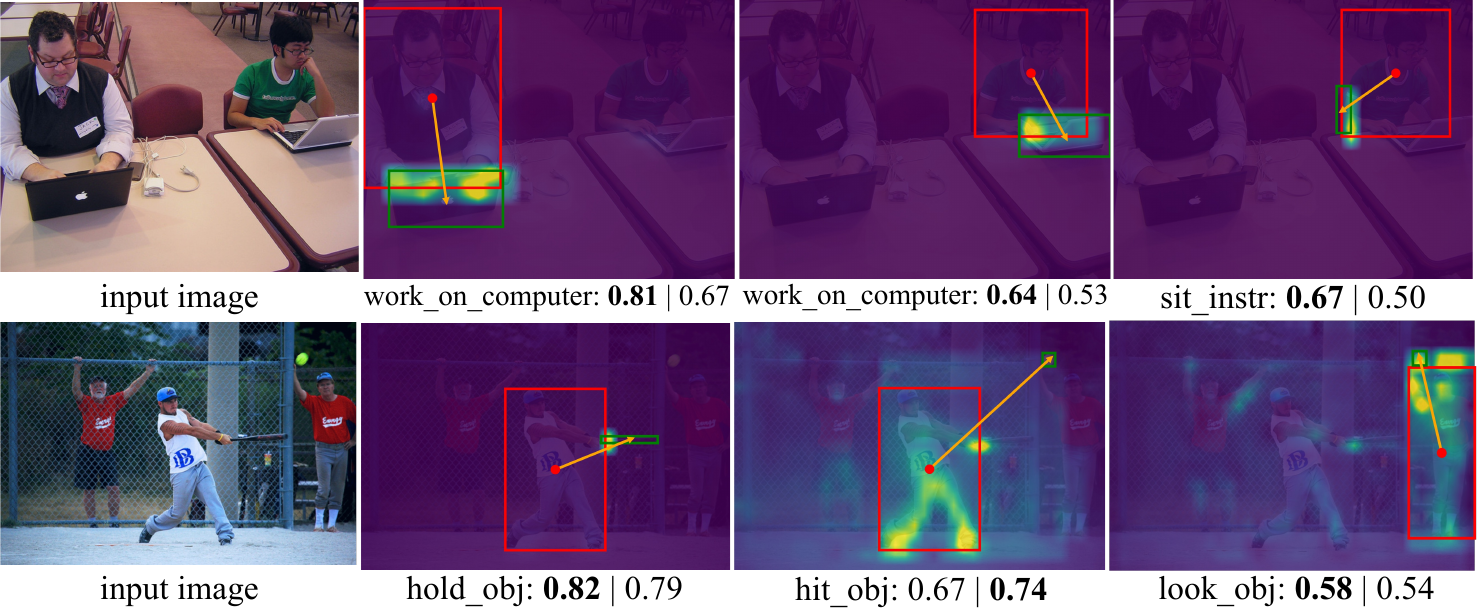}
\caption{\footnotesize{Visualization of the attention in simple cases. The prediction scores of GeoHOI and STIP are shown (left: GeoHOI, right: STIP).}}
\label{Fig: attention_vis}
\end{figure}

\begin{figure}[htbp]
\setlength{\abovecaptionskip}{1pt} 
\centering
\includegraphics[width=\linewidth]{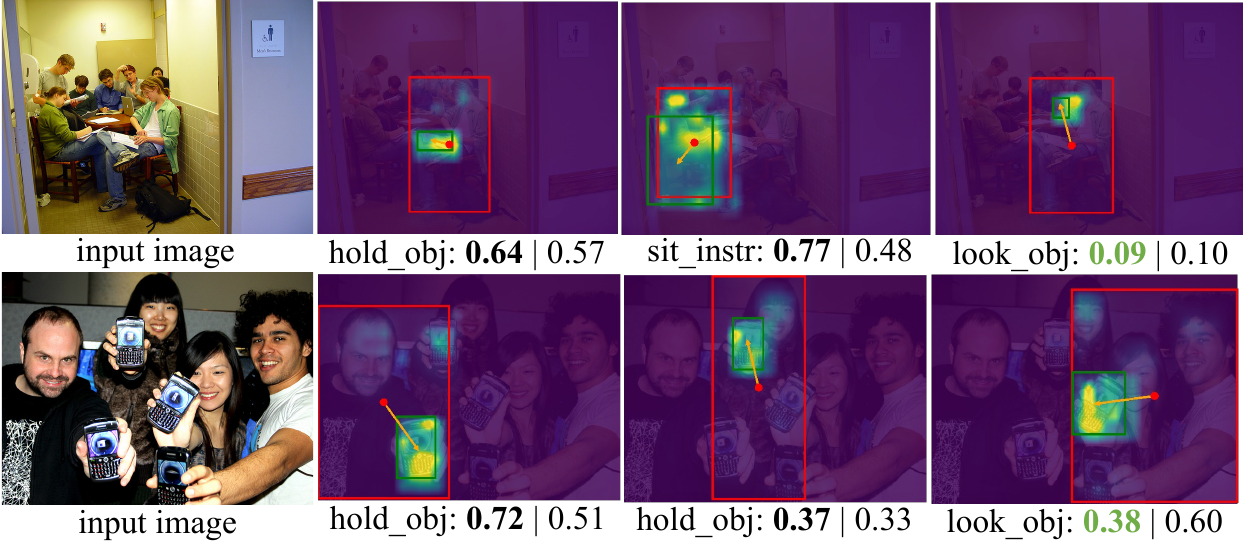}
\caption{\footnotesize{Visualization of the attention in complex cases. The prediction scores of GeoHOI and STIP are shown (left: GeoHOI, right: STIP). The black color represents true positive interactions and the green color means true negative interactions.}}
\label{Fig: attention_vis_add}
\end{figure}

In Table \ref{table: perclass}, we report the per-class performance of GeoHOI and compare it with the backbone model STIP on V-COCO. We run the pre-trained checkpoint of STIP to obtain its per-class results since they are not provided in the original paper. We can see that GeoHOI outperforms the backbone STIP in the majority of classes, particularly in the ``eat-obj", ``lay-instr", ``carry-obj", and ``drink-instr" classes. From the upper row of Fig. \ref{Fig: case-study}, objects in these classes are generally either partially occluded with humans or quite large. For example, the objects in ``drink-instr" are often occluded with human hands. In these cases, we believe that keypoints can provide valuable information on the visible parts of these objects and how they are being interacted with the human, resulting in enhanced performance. Moreover, objects such as beds and surfboards, often associated with the ``lay-instr" and ``carry-obj" actions, are typically quite large. The detected keypoints can capture their shapes pretty well. As a result, it boosts the performance of interaction detection.

On the other hand, GeoHOI performs much worse than STIP in the ``hit-instr", ``hit-obj", and ``talk\_on\_phone" classes. First, objects such as baseball bats and tennis rackets shown in the lower row of Fig. \ref{Fig: case-study}, typically appear in crowded scenes, leading to inaccurate object detection. Second, the inherent slender shape of baseball bats and the varying perspectives of tennis rackets hinder our UniPointNet from detecting their representative keypoints effectively. Third, balls associated with the ``hit-obj" action and cell phones in the ``talk\_on\_phone" action are too small. Thus, their masks have very small areas compared to the entire images, leading to noisy keypoints that harm interaction detection.

Fig. \ref{Fig: attention_vis} shows qualitative results and compares GeoHOI with the backbone STIP. The top 3 interaction prediction probabilities of GeoHOI are visualized. The images show the variance in object sizes, human visibilities, and different interaction classes. First, the attention maps highlight different local regions for the same interaction category in the same image. For example, the hands are highlighted in different regions for action ``work\_on\_computer'' as shown in the first row. Second, when the human and object are far away from each other, they also gather a certain amount of information from their neighbourhood as illustrated in the second row, indicating both local and cross-instance cues are essential for HOI classification. We further showcase crowded scenes with multiple humans and objects in Fig. \ref{Fig: attention_vis_add}. GeoHOI shows higher confidence for true interaction actions ``hold\_obj'' and ``sit\_instr'' and less confidence for true negative interaction ``look\_obj'', indicating its effectiveness. Overall, GeoHOI improves the backbone of STIP by predicting higher scores in various true interactions and lower scores in negative ones in most cases. More qualitative results of GeoHOI on complex scenes can be found in the supplementary material.

\begin{table}[htbp]
    \scriptsize
    \centering
    \caption{Comparison of performance, number of parameters, and time for training and inference on V-COCO. We conduct both the training and inference process with a batch size of 1 on a single Quadro RTX 5000 GPU. The number of parameters is represented in millions (M). The Speed means the elapsed time (ms) for processing one image.}
    \label{table: complexity}
    \begin{tabular}{cccccc}\hline
    Method&${AP}_{role}^{\#1}\uparrow$&${AP}_{role}^{\#2}\uparrow$&\#Params$\downarrow$&Training$\downarrow$&Inference$\downarrow$ \\ \hline \hline
    STIP \cite{stip} & 66.0 & 70.7 & 13.2 M & 143.2 ms & 117.8 ms \\ 
    GeoHOI& 67.8 & 73.3 & 17.8 M & 794.4 ms & 588.6 ms \\ 
    \hline		
    \end{tabular}
\end{table}

In Table \ref{table: complexity}, we report the performance, number of parameters, and speed for training and inference on V-COCO of STIP and GeoHOI for an objective comparison. GeoHOI outperforms STIP by a significant margin of 1.8 and 2.6 in terms of mAP in ${AP}_{role}^{\#1}$ and ${AP}_{role}^{\#2}$ with a comparable number of parameters. GeoHOI's training and inference time is slower than STIP but remains within an acceptable one-second threshold. Compared to STIP, GeoHOI requires two additional modules (object segmentation and keypoint detection) executed sequentially, resulting in higher time consumption.

\subsection{Ablation Studies}
In this section, we analyze each GeoHOI design by discussing its possible variants on V-COCO to provide more insights. All experiments are carried out under the training setting of the pre-trained object detector with frozen weights. 

\begin{table}[htbp]
    \scriptsize
    \centering
    \caption{Performance contribution analysis of each component in GeoHOI on V-COCO.}
    \label{table: component_analysis}
    \begin{tabular}{l c c}\hline
    Method & ${AP}_{role}^{\#1}$ & ${AP}_{role}^{\#2}$ \\ \hline \hline
    Baseline & 65.2 & 69.8 \\ \hline
    + KIP & 66.2 ($\uparrow 1.0$) & 71.0 ($\uparrow 1.2$) \\ 
    + PAM (w/o human patch) & 66.2 ($\uparrow 1.0$) & 71.1 ($\uparrow 1.3$) \\ 
    + PAM (w/o object patch) & 66.5 ($\uparrow 1.3$) & 71.4 ($\uparrow 1.6$) \\ 
    + PAM (w/o positional encodings) & 66.3 ($\uparrow 1.1$)  & 71.1 ($\uparrow 1.3$) \\ 
    + PAM & 66.7 ($\uparrow 1.5$) & 71.9 ($\uparrow 2.1$) \\ 
    + KIP + PAM (GeoHOI) & \textbf{67.8} ($\uparrow 2.6$) & \textbf{73.3} ($\uparrow 3.5$) \\ 
    \hline
		
    \end{tabular}
\end{table}

\textbf{Impact of Individual Components.} We conduct ablation experiments by comparing different variants of GeoHOI in Table \ref{table: component_analysis}. We start with the baseline model (\textbf{Baseline}), which adopts the structure-aware HOI network introduced in \cite{stip}, but with its object detector replaced by Panoptic DETR. Next, we extend the Baseline model by integrating our keypoint-aware graph convolution network into its interactiveness prediction module, incorporating the holistic graph features, yielding \textbf{Baseline + KIP} which demonstrates better performance. After that, we enhance the Baseline model with our Part Attention Module but without human patches. This variant of our model (\textbf{Baseline + PAM (w/o human patch)}) achieves better performance than both the Baseline model and Baseline + KIP. Another variant of our model (\textbf{Baseline + PAM (w/o object patch)}) shows even better performance. This indicates that the human patch features are more important than the object patch features. We believe this is because the rich poses of humans captured by keypoints are more beneficial for recognizing interactions, which aligns with the findings in \cite{SGCN4HOI}. To evaluate the benefits of using keypoints as positional encodings, we create (\textbf{Baseline + PAM (w/o positional encodings)}). It outperforms other variants, though it is slightly worse than \textbf{Baseline + PAM} that incorporates both patch features and keypoints. Both Baseline + PAM (w/o positional encodings) and Baseline + PAM demonstrate the effectiveness of using local features with self-attention. Finally, when jointly upgrading the Baseline model with the keypoint-aware interactiveness prediction module and part attention module (i.e., our \textbf{GeoHOI}), it results in the best performance.

\begin{table}[htbp]
\scriptsize
    \centering
    \caption{Performance comparison by using the different number of keypoints ($N$) on V-COCO.}
    \label{table: AS_num_kp}
    \begin{tabular}{c c c}\hline
    \# of keypoints ($N$) & ${AP}_{role}^{\#1}$ & ${AP}_{role}^{\#2}$ \\ \hline \hline
    4 & 65.6 & 70.3 \\ 
    8 & 66.6 & 71.5 \\ 
    16 & 66.8 & 71.7 \\ 
    \textbf{32} & \textbf{67.8} & \textbf{73.3} \\ 
    48 & 67.0 & 71.9 \\ 
    \hline
		
    \end{tabular}
\end{table}

\begin{figure}[htbp]
\setlength{\abovecaptionskip}{1pt} 
\centering
\includegraphics[width=0.8\linewidth]{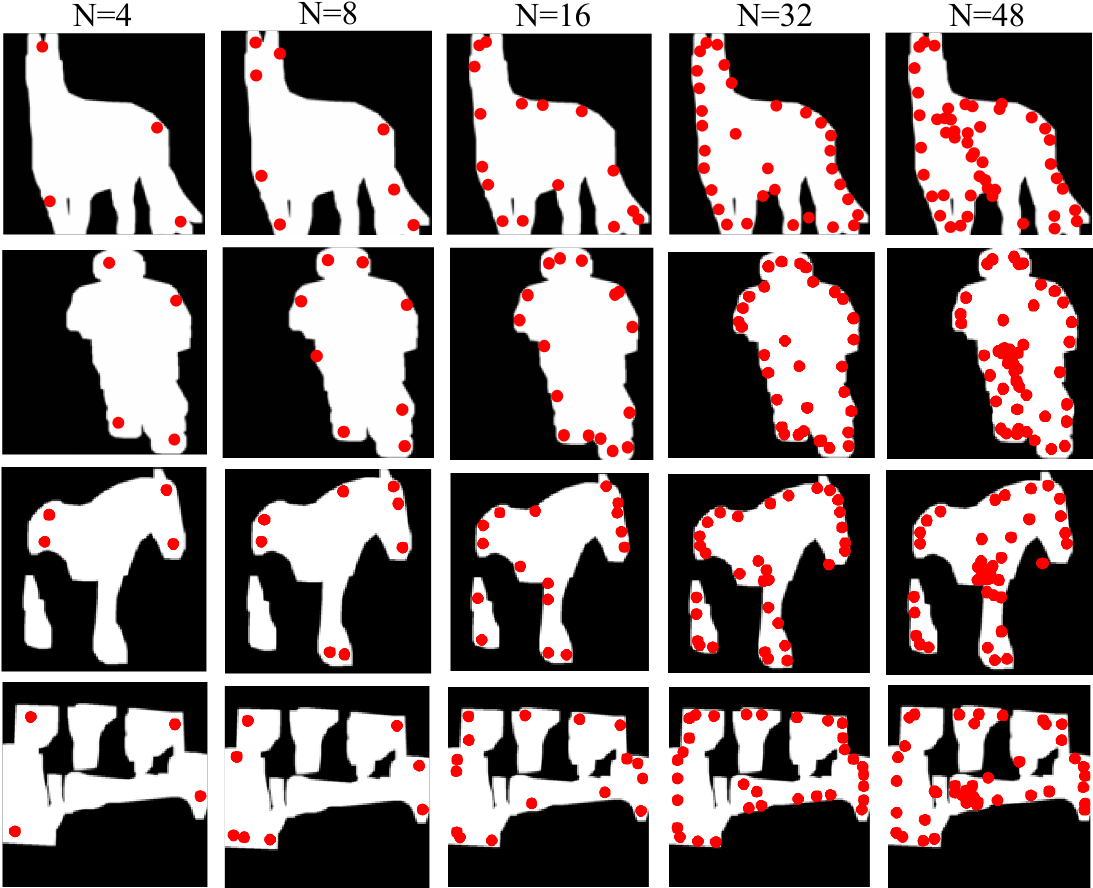}
\caption{\footnotesize{Examples illustrating different numbers of keypoints. When there are very few keypoints, they only represent the basic shape of an object. For example, with $N=4$, only edge corners are represented. With abundant keypoints, the representation can be redundant.}}
\label{Fig: AS_num_kp}
\end{figure}

\textbf{Effect of Different Numbers of Keypoints.} Here, we vary $N$ from 4 to 48 to demonstrate the relationship between the performance and the select keypoints number $N$. Table \ref{table: AS_num_kp} shows the quantitative ablation tests, and the best performance is obtained when $N$ is 32. The increasing number of keypoints (until $N=32$) can generally boost the performance. This is expected since more details can be captured with more keypoints. For example, when $N=4$, only the upper part of the horse is modelled as shown in the first column of the third row in Fig. \ref{Fig: AS_num_kp}. In addition, when the object's mask is separated into multiple fragments due to occlusion (as seen in the last two rows in Fig. \ref{Fig: AS_num_kp}), a higher number of keypoints can span more of these segments, generating a more accurate shape representation of an object. However, when $N$ is greater than 32, i.e., 48, the performance decreases. We speculate that too many keypoints might introduce more noise, leading the model to overfit, which results in affecting its generalization. As such, we have empirically selected $N$ to be 32.

\begin{table}[htbp]
    \scriptsize
    \centering
    \caption{Effect of UniPointNet in GeoHOI on V-COCO.}
    \label{table: effect_unipointnet}
    \begin{tabular}{c c c}\hline
    Method & ${AP}_{role}^{\#1}$ & ${AP}_{role}^{\#2}$ \\ \hline \hline
    GeoHOI (UniPointNet) & \textbf{67.8} & \textbf{73.3} \\ 
    GeoHOI (Skeletal Keypoint \cite{SGCN4HOI}) & 66.7 & 71.3 \\ 
    \hline
		
    \end{tabular}
\end{table}

To evaluate the effectiveness of UniPointNet, in Table \ref{table: effect_unipointnet}, we compare it with the existing skeleton-based keypoint representation (Skeletal Keypoint) for HOI detection \cite{SGCN4HOI}, which also utilizes object segmentation. GeoHOI (UniPointNet) means the keypoints in GeoHOI are detected by our proposed UniPointNet, and GeoHOI (Skeletal Keypoint) represents that the keypoints are obtained from \cite{SGCN4HOI}. We can see that UniPointNet surpasses the skeletal keypoints in both ${AP}_{role}^{\#1}$ and ${AP}_{role}^{\#2}$, showcasing the effectiveness of the proposed UniPointNet. The Skeletal Keypoint is a skeleton-driven method, it is only robust for articulated objects like humans and dogs, which demonstrate a clear and consistent structure of joints and parts. It is also limited when extracting keypoints from non-articulated objects such as pizzas and phones because of its skeletonization process. In contrast, UniPointNet is a shape-driven representation, making it robust to arbitrary shapes of objects.

\begin{figure*}[htbp]
\setlength{\abovecaptionskip}{1pt} 
\centering
\includegraphics[width=0.95\linewidth]{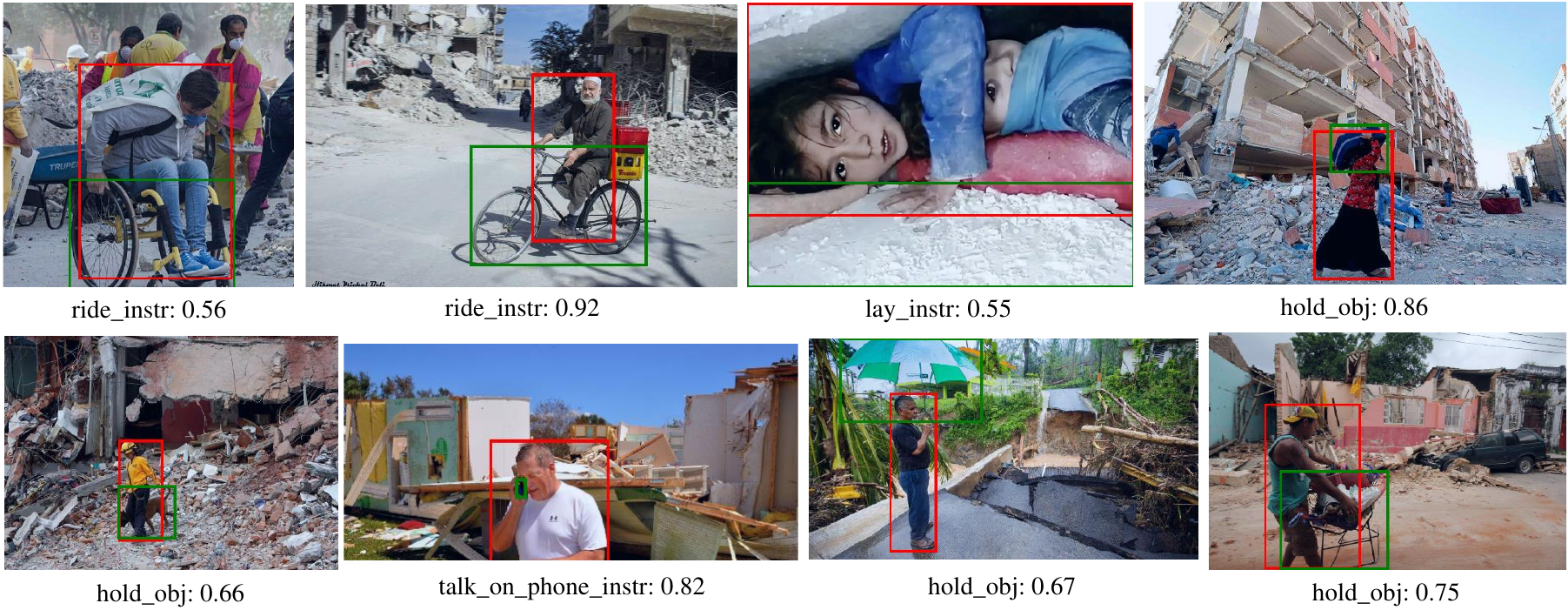}
\caption{\footnotesize{Case study in post-disaster rescue. Different interactions and scenes are shown, and the top 1 interaction of each image is given.}}
\label{Fig: application_case}
\end{figure*}

\subsection{A Case Study in Post-Disaster Rescue with UAVs}

The proposed HOI detector (GeoHOI) has a wide range of applications, including vision-based instrumentation and measurement. To showcase the generalization of our GeoHOI and evaluate its performance in real-world applications relevant to instrumentation and measurement, we conducted a case study in Post-Disaster Rescue with unmanned aerial vehicles (UAVs) on the PDD dataset \cite{post-disaster}. It was collected from real-world ruins, including various post-disaster scenes, from multiple angles of UAVs and different distances and resolutions. Disasters include natural calamities such as earthquakes and outdoor rescue scenarios, among others. It consists of 832 training, 100 validation, and 100 testing images.

By conducting the experiments on the PDD dataset, we evaluate our GeoHOI for human detection task and compare it with the baseline methods. GeoHOI is designed for HOI detection, outputting triplets as $\left<human, interaction, object \right>$. To evaluate it on human detection, we measure its outputs of detected human bounding boxes and ignore interaction and object bounding box predictions. To make a fair comparison between our methods and baselines in \cite{post-disaster}, we requested the PDD test set (100 images) from the authors for evaluation, and we also used the same evaluation metrics, i.e., average precision (AP), F1 score, recall, and precision. We directly apply the pre-trained GeoHOI and STIP (both are trained on V-COCO) on the test set of the PDD dataset. In addition, we conduct a qualitative analysis of HOI detection to showcase that the proposed method can further facilitate post-disaster rescue with UAVs. For example, detecting individuals in wheelchairs or needing medical assistance allows rescue teams to effectively prioritize rescue efforts such as aid and resources for those who need them most.

\begin{table}[htbp]
    \centering
    \caption{Performance evaluation of generalizing GeoHOI in human detection on the PDD test set. ``Detection-based'' denotes human detection models used in the PDD dataset, and ``HOI-based'' represents HOI detection models that are evaluated on human detection. Note that ``$K$'' is the number of output proposals from our KIP module.}
    \label{table: performance_pdd}
    \resizebox{\linewidth}{!}{
    \begin{tabular}{c c c c c c}\hline
    & Model & AP@0.5 & F1 & Recall & Precision \\ \hline
    \multirow{19}{*}{Detection-based \cite{post-disaster}}& YOLOv5s & 84.15\% & 0.86 & 80.31\% & 93.58\% \\
    & YOLOv5m & 84.36\% & 0.87 & 82.49\% & 92.58\%  \\
    & YOLOv5l & 84.25\% & 0.88 & 83.92\% & 93.45\%  \\ 
    & im-YOLOv5s & 80.82\% & 0.85 & 78.43\% & 92.59\% \\
    & im-YOLOv5m & 83.32\% & 0.88 & 82.28\% & 94.14\%  \\
    & im-YOLOv5l & 84.38\% & 0.87 & 82.03\% & 93.33\%  \\ 
    & YOLOv7 & 85\% & 0.86 & 82.56\% & 90.64\%  \\
    & YOLOv7x & 87.35\% & 0.89 & 85.88\% & 91.63\%  \\ 
    & YOLOv8s & 85.52\% & 0.85 & 83.33\% & 87.76\% \\
    & YOLOv8m & 90.78\% & 0.89 & 85.66\% & 91.70\%  \\
    & YOLOv8l & 87.81\% & 0.88 & 84.05\% & 91.53\%  \\ 
    & YOLO-NASs & 86.08\% & 0.86 & 86.61\% & 85.27\% \\
    & YOLO-NASm & 85.94\% & 0.86 & 86.38\% & 85.06\%  \\
    & YOLO-NASl & 87.26\% & 0.86 & 88.72\% & 84.13\%  \\ 
    & DETR & 87.89\% & 0.88 & 89.08\% & 87.24\% \\
    & DDETR & 76.87\% & 0.64 & 84.23\% & 51.01\%  \\
    & DAB-DETR & 88.78\% & \textbf{0.91} & 88.42\% & 93.47\% \\
    & DN-DETR & 87.17\% & 0.9 & 86.15\% & \textbf{94.92}\%  \\
    & DINO & 91.03\% & \textbf{0.91} & 89.66\% & 92.86\%  \\ 
    & faster R-CNN & 88.52\% & 0.87 & 89.19\% & 84.62\%  \\ \hline
    \multirow{4}{*}{HOI-based} & STIP & 91.01\% & 0.80 & 71.51\% & 91.58\% \\
    & GeoHOI ($K = 32$) & \textbf{92.32\%} & 0.81 & 71.94\% & 92.63\%  \\ 
    & GeoHOI ($K = 64$) & 88.65\% & 0.88 & 89.41\% & 88.04\%  \\ 
    & GeoHOI ($K = 100$) & 84.64\% & 0.88 & \textbf{91.48}\% & 86.49\%  \\ \hline
		
    \end{tabular}}
\end{table}

In Table \ref{table: performance_pdd}, we compare the quantitative performance of HOI-based models, including our proposed GeoHOI and its backbone STIP and baselines proposed in \cite{post-disaster}. With the default number of 32 output proposals in the Keypoint-aware Interactiveness Prediction (KIP) module, GeoHOI outperforms all the baselines on AP and achieves comparable precision, demonstrating its effectiveness in detecting humans in post-disaster scenes. STIP obtains similar performance in both AP and precision, and we believe the main reason is that the HOI-based detection systems can enhance human bounding box precision by leveraging contextual information (i.e., interactions between humans and objects) and joint optimization (i.e., optimizing the predictions of humans, interactions, and objects simultaneously). The integrated analysis of humans, objects, and their interactions refines human detection accuracy compared to these baselines designed alone for human detection. The lower performance on the F1 score and recall of GeoHOI and STIP indicate that the HOI-based systems have a higher missed detection rate. We think the KIP module that suppresses non-interactive human-object pairs is the primary cause since it can filter out humans who do not interact with objects, resulting in compromised performance in recall. To verify this, we increase the number of proposals ($K$) to 64 and 100, respectively. Recall significantly improves with the number of proposals and outperforms all the baselines at $K = 100$. This indicates our model's adaptability in balancing recall and precision by tuning the number of output proposals in our KIP module in practical applications.

In addition, we show the qualitative results of HOI detection in Fig. \ref{Fig: application_case} to provide an in-depth analysis of how HOI detection can facilitate post-disaster rescue. GeoHOI demonstrates a varied performance across different scenarios. For instance, it predicts relatively high confidence scores in recognizing the interactions of ``ride\_instr'' (riding a bicycle), ``talk\_on\_phone\_instr'', and ``hold\_obj'' where the scenes are less complicated. In contrast, it shows diminished confidence in more complex scenes, such as a person lying down in the rubble, or when the scene is crowded, e.g., the image in the first row and column. This indicates the challenges in detecting interactions in cluttered post-disaster scenes.

The qualitative results show that our proposed GeoHOI is able to detect different human interactions in post-disaster scenes, facilitating search and rescue operations. For instance, identifying individuals in wheelchairs or those lying on the ground enables rescue teams to prioritize medical attention. Observations of people using phones or riding bicycles provide crucial insights into the operational status of communication networks and the accessibility of various areas. Additionally, recognizing survivors holding onto pets or personal belongings allows rescue teams to provide not only necessities like food and water but also support for pet care and the safekeeping of valuables, enhancing the overall rescue operation.

\section{Conclusion}
In this paper, we have proposed GeoHOI, an end-to-end Transformer-style model for detecting human-object interactions using fine-grained geometric keypoint features of humans and objects. We have also presented UniPointNet, a self-supervised framework that detects keypoints for arbitrary objects and enhances HOI performance. The KIP module uses keypoints to mine cross-instance cues via a graph network, enhancing pairwise cues for optimizing the prediction of interactive human-object pairs. The PAM module uses self-attention on keypoint patches to discover informative local cues, facilitating the prediction of specific interaction categories. Extensive experimental results have shown that GeoHOI improves the backbone of STIP and achieves superior performance on public HOI benchmarks. We further demonstrated the advantages of using GeoHOI on human-centric applications such as the case study on post-disaster rescue. The presented UniPointNet also facilitates visual measurement tasks, including object pose estimation \cite{object-pose} and 3-D reconstruction \cite{reconstruction}.

The end-to-end GeoHOI is limited in training and analysis of the geometric features. For future research, the proposed geometric features can be employed in two-stage frameworks such as \cite{viplo}, facilitating more analysis and insights into the geometric context (e.g., relative keypoint distance) in HOI detection. As discussed in the experiments, our UniPointNet struggles with tiny or slender objects due to their limited spatial resolution in images, which hinders accurate shape reconstruction and keypoint detection. Future work could explore better keypoint representation for these objects, such as adaptively selecting the optimal numbers and locations of keypoints to represent objects in different sizes. Additionally, investigating how to incorporate semantic information in keypoint detection and evaluating the effect on HOI detection would also be valuable. 

Furthermore, recent advancements in large language models, especially those with integrated vision-language capabilities such as CLIP \cite{CLIP}, have demonstrated their effectiveness in zero-shot HOI detection \cite{hoiclip,SQA}. Given that annotating HOI triplets is challenging and rare HOIs are not learned as effectively as non-rare ones, it is worth further exploring the capabilities of large language models in the future to tackle the long-tail problem and zero-shot learning in HOI detection, facilitating real-world HOI applications.

\section*{Acknowledgement}
This research is supported in part by the EPSRC NortHFutures project (ref: EP/X031012/1).

{\tiny
\bibliographystyle{IEEEtran}  
\bibliography{IEEEabrv,references}
}

\newpage

\title{Geometric Features Enhanced Human-Object Interaction Detection (\textit{Supplementary Material)}}
\author{Manli Zhu$^{\orcidlink{0000-0002-8231-5342}}$, Edmond S. L. Ho$^{\orcidlink{0000-0001-5862-106X}}$, Shuang Chen$^{\orcidlink{0000-0002-6879-7285}}$, Longzhi Yang$^{\orcidlink{0000-0003-2115-4909}}$, \textit{Senior Member, IEEE}, Hubert P. H. Shum$^{\orcidlink{0000-0001-5651-6039}\dag}$, \textit{Senior Member, IEEE}
}


\markboth{IEEE TRANSACTIONS ON INSTRUMENTATION AND MEASUREMENT}
{}

\maketitle

\section{Supplementary Material}
Here we show more qualitative results of GeoHOI on complex scenes to showcase its effectiveness and robustness.

\begin{figure*}[hb]
\setlength{\abovecaptionskip}{1pt} 
\centering
\includegraphics[width=\textwidth]{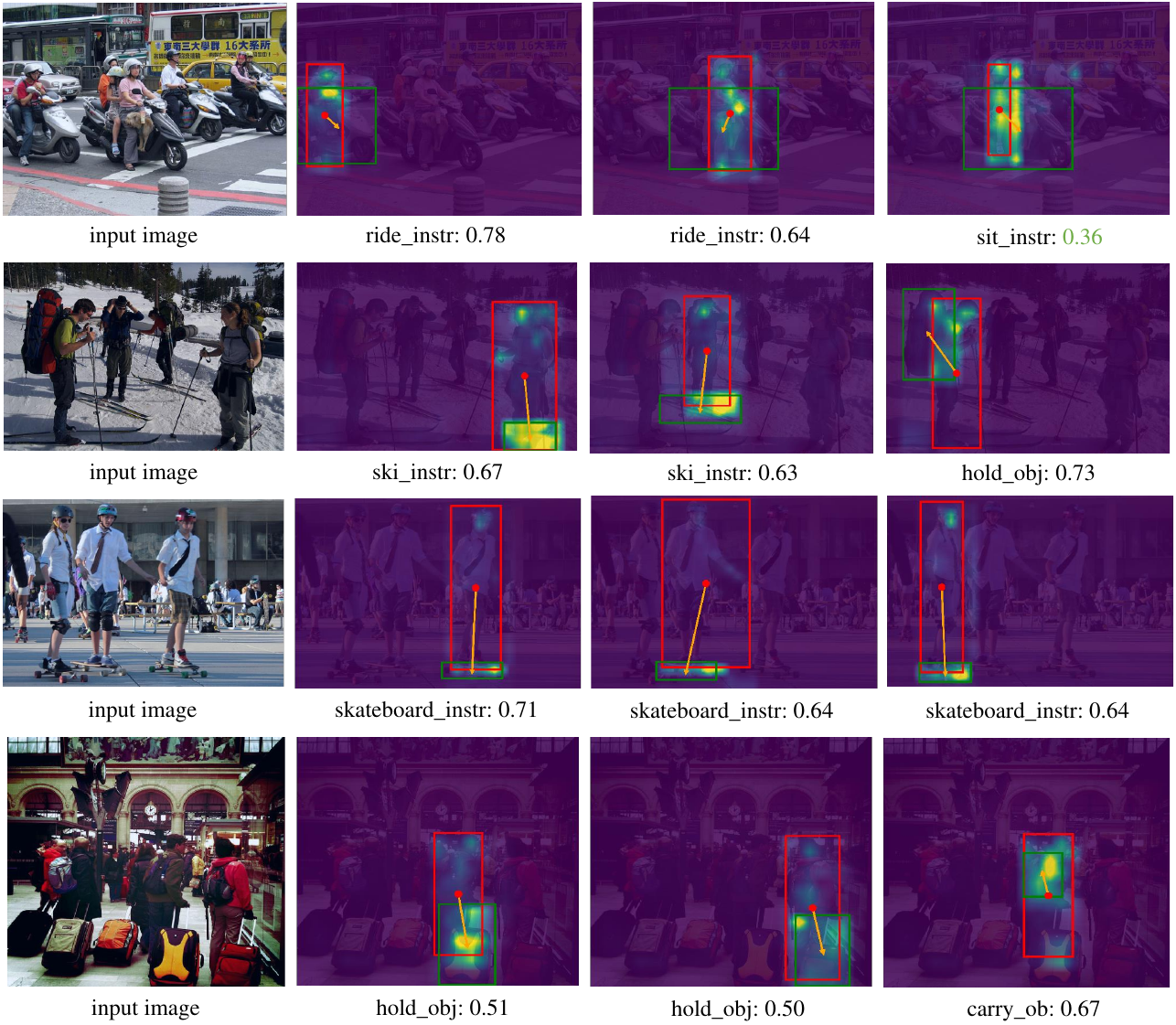}
\caption{\footnotesize{Qualitative results of GeoHOI. The black prediction scores represent true positive interactions and the green ones indicate true negative interactions.}}
\label{Fig: more_attention_1}
\end{figure*}

\begin{figure*}[htbp]
\setlength{\abovecaptionskip}{1pt} 
\centering
\includegraphics[width=\textwidth]{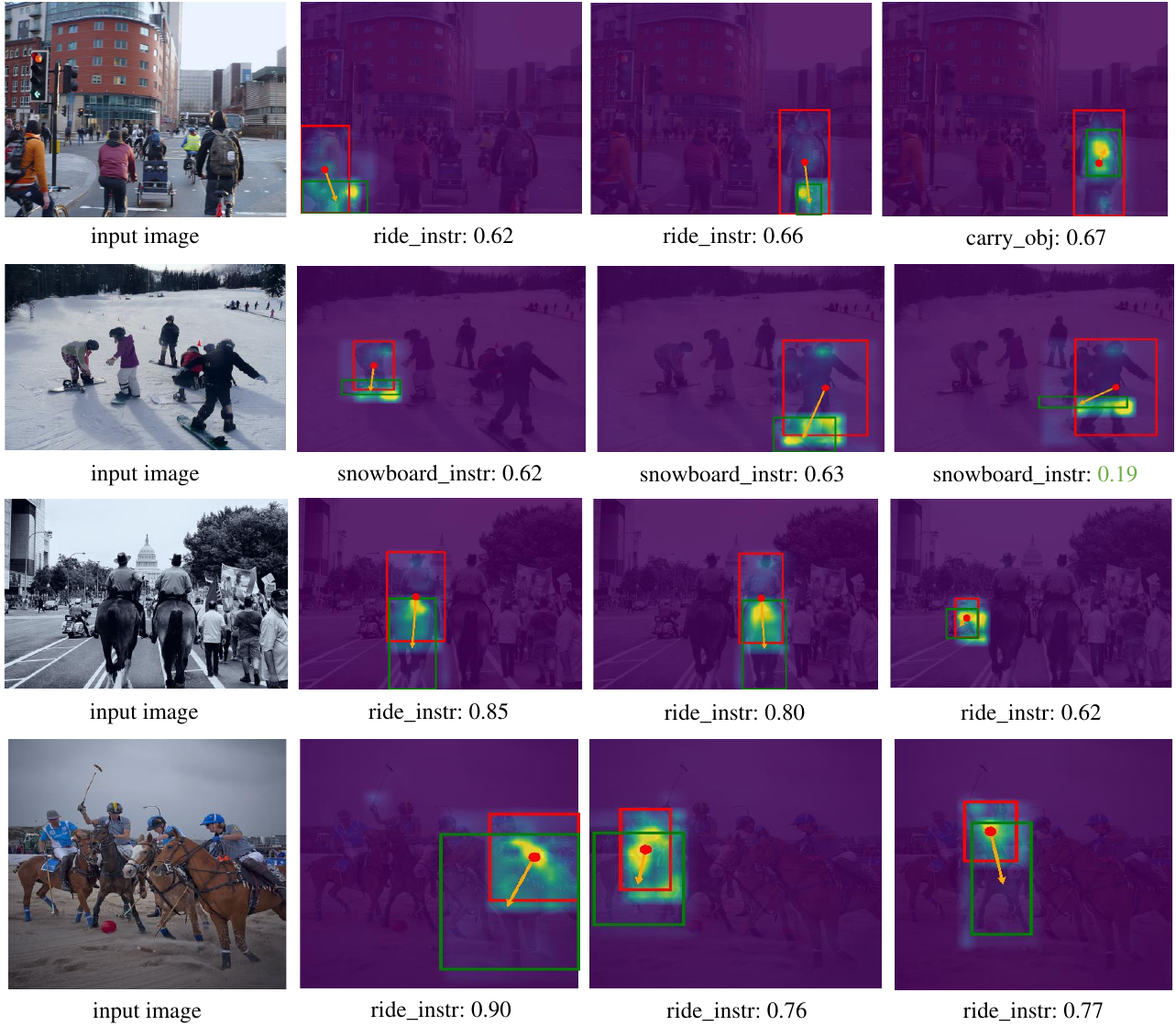}
\caption{\footnotesize{Qualitative results of GeoHOI. The black prediction scores represent true positive interactions and the green ones indicate true negative interactions.}}
\label{Fig: more_attention_2}
\end{figure*}

\end{document}